\documentclass{article} % For LaTeX2e
\usepackage{iclr2027_conference,times}

\usepackage{amsmath,amsfonts,bm}

\def\eqref#1{equation~\ref{#1}}
\def\1{\bm{1}}

\def\rmI{{\mathbf{I}}}

\def\rmS{{\mathbf{S}}}

\def\vb{{\bm{b}}}
\def\vc{{\bm{c}}}

\def\vk{{\bm{k}}}

\def\vo{{\bm{o}}}

\def\vq{{\bm{q}}}

\def\vv{{\bm{v}}}
\def\vw{{\bm{w}}}

\DeclareMathAlphabet{\mathsfit}{\encodingdefault}{\sfdefault}{m}{sl}
\SetMathAlphabet{\mathsfit}{bold}{\encodingdefault}{\sfdefault}{bx}{n}

\usepackage{hyperref}
\usepackage{url}
\usepackage{amsmath}
\usepackage{amssymb}
\usepackage{amsfonts}            % blackboard math symbols
\usepackage{amsthm}
\usepackage{mathtools}
\usepackage{threeparttable}
\usepackage{multirow}
\usepackage{bm}
\usepackage{bbm}
\usepackage{nicefrac}            % compact symbols for 1/2, etc.

\usepackage[table]{xcolor}       % colours (table option pulls in colortbl)
\usepackage{booktabs}            % professional-quality tables
\usepackage{multirow}
\usepackage{multicol}
\usepackage{cleveref}
\usepackage{graphicx}
\usepackage{wrapfig}
\usepackage{float}
\usepackage{tikz}
\usepackage{pgfplots}
\pgfplotsset{compat=1.18}
\usepgfplotslibrary{fillbetween}
\usepgflibrary{patterns}
\usetikzlibrary{arrows.meta,positioning,calc,shapes.geometric,decorations.text,
  patterns,patterns.meta,fit,backgrounds,chains,shadows,math,circuits.ee.IEC,
  decorations.pathmorphing,decorations.pathreplacing,decorations.shapes}

\usepackage[labelfont=bf,font=small]{caption}
\usepackage{subcaption}          % replaces the deprecated 'subfigure'
\usepackage{listings}
\usepackage{minted}              % NOTE: requires pdflatex -shell-escape + Pygments

\usepackage{url}
\usepackage{textcomp}
\usepackage{pifont}
\usepackage{scalefnt}
\usepackage{tcolorbox}
\tcbuselibrary{skins}
\usepackage{fontawesome}
\usepackage{lipsum}              % 

\usepackage{hyperref}

\usepackage{booktabs}
\usepackage{adjustbox}
\usepackage{multirow}
\usepackage[table]{xcolor}
\usepackage{colortbl}
\usepackage{wrapfig}
\usepackage{booktabs}
\usepackage{graphicx}
\usepackage{caption}
\usepackage[table]{xcolor}

\usepackage{threeparttable}

\newcommand{\rulerbest}[1]{\textbf{#1}}
\newcommand{\rulersecond}[1]{\underline{#1}}
\newcommand{\rulermodel}[1]{\hspace{1.2mm}#1}

\definecolor{codebg}{RGB}{245,245,245}
\definecolor{keywordcolor}{RGB}{0,0,153}
\definecolor{commentcolor}{RGB}{34,139,34}
\definecolor{stringcolor}{RGB}{163,21,21}
\definecolor{numbercolor}{RGB}{128,128,128}
\title{Gated Slot Attention-2: Two-Sided Associative Memory
Correction in Linear Attention}

\author{
\textbf{
Ruijie Li$^{1}$,
Shengnan Ding$^{2}$,
Weimin Zhang$^{2,*}$,
Derick Tang$^{2,*}$,
Zhanpeng Zeng$^{2}$
} \\
\textbf{
Qinsong Zeng$^{2}$,
Ming Chen$^{2}$,
Jiaxi Hu$^{1,*}$,
Yuxuan Liang$^{1,*}$
} \\
\normalfont
$^{1}$The Hong Kong University of Science and Technology (Guangzhou) \\
$^{2}$Tencent \\
$^{*}$Corresponding authors
}

\iclrfinalcopy

\begin{document}

\maketitle

\begin{abstract}

Linear attention models have emerged as efficient alternatives to
standard attention, but effectively managing their fixed-size recurrent
memory remains challenging.
To improve memory, recent work has explored two distinct directions:
delta-rule variants for precise correction of values associated with
keys, and slot-based architectures such as Gated Slot Attention
for modeling key and value memories in two stages.
We observe that these directions are complementary--the delta rule
provides effective memory correction, while the two-stage structure
provides a natural way to operate on both sides of an association.
Building on this insight, we introduce a new Gated Oja Rule\footnote{
Oja's rule is a normalized Hebbian learning rule that augments Hebbian
association with a self-normalizing correction term
~\citep{oja1982simplified}; our gated Oja rule adapts this structure
to recurrent associative memory.
} for key-side correction and extend it with decoupled erase and write
control to obtain Gated Oja Rule-2.
We then introduce Gated Slot Attention-2 (GSA2), which combines
Gated Oja Rule-2 for key-side correction with Gated Delta Rule-2 for
value-side correction through shared latent slots.
We further derive a hardware-efficient chunkwise algorithm for parallel training.
Experiments demonstrate that GSA2 consistently improves over strong
linear-attention baselines across benchmarks while retaining linear-time
sequence modeling and constant-memory recurrent decoding.

\end{abstract}

\section{Introduction}
\label{sec_introduction}
The Transformer~\citep{vaswani2017attention} has become the dominant architecture for large language models and sequence modeling due to the strong expressiveness of self-attention and its natural compatibility with parallel training. 
However, the computational cost of self-attention grows quadratically with
sequence length, while the key-value cache grows linearly during
autoregressive decoding. 
Linear attention~\citep{katharopoulos2020transformers} offers an appealing alternative by compressing the entire history into a fixed-size recurrent state, enabling linear-time sequence modeling and constant-memory recurrent inference. 
The central challenge, however, is how to effectively organize and update information within this compressed memory.

Recent progress on linear attention has improved fixed-state memory through data-dependent decay, selective gating, and increasingly expressive state transitions~\citep{sun2023retentive,yang2023gated,gu2023mamba,qin2024hgrn2,dao2024transformers,hu2025comba,lahoti2026mamba}. Within this landscape, Gated Slot Attention (GSA)~\citep{zhang2024gated} stands out as an elegant variant: rather than refining the update of a single recurrent state, it maintains separate key and value memories over a shared set of latent slots. This slot-based factorization provides a natural interface for operating on both sides of an association while retaining a compact recurrent memory.

In parallel, another line of work has focused on improving the memory update
itself, with delta-rule models providing an effective mechanism for explicit
associative correction.
DeltaNet~\citep{yang2024parallelizing} corrects the value associated with
the current key, Gated DeltaNet (GDN)~\citep{yang2025gated} adds adaptive
forgetting, Kimi Delta Attention (KDA)~\citep{team2025kimi} extends decay
to the channel level, and Gated DeltaNet-2 (GDN2)~\citep{hatamizadeh2026gated}
further decouples erasing old content from writing new content.
Despite these advances, this line still centers on a single key-addressed
associative state whose correction primarily acts on the value side.

We observe that these two directions are complementary.
GSA provides an explicit two-stage structure spanning both sides, but lacks similarly expressive correction rules;
Delta-rule models provide powerful error-correcting memory updates, but primarily correct the value side of an association.
This suggests a simple question:
\begin{tcolorbox}[enhanced, width=0.95\linewidth, center,
    colback=white, colframe=gray!60!black, boxrule=1pt,
    arc=2mm, drop fuzzy shadow]
    \centering \itshape
    Can we equip both stages of a slot-based associative memory with complementary error-correcting updates?
\end{tcolorbox}

Building on this insight, we first introduce the \textbf{Gated Oja Rule (OJA)},
inspired by Oja's normalized Hebbian update~\citep{oja1982simplified},
to provide the complementary key-side correction.
We then decouple its erase and write operations, yielding
\textbf{Gated Oja Rule-2 (OJA2)}.
Finally, we introduce \textbf{Gated Slot Attention-2 (GSA2)}, which
combines Gated Oja Rule-2 with the Gated Delta Rule-2 of GDN2 through
shared latent slots.
The first stage associates keys with latent slots and applies Gated Oja
Rule-2 to correct the key side of the association, while the second
associates the same slots with values and applies Gated Delta Rule-2 to
correct the value side.

Let $\bm{k}_t$, $\bm{w}_t$, and $\bm{v}_t$ denote the key, shared
slot-space vector, and value at step $t$, respectively.
GSA2 can then be summarized as
\[
\bm{k}_t
\xleftarrow[\text{key-side correction}]{\text{Gated Oja Rule-2}}
\bm{w}_t
\xrightarrow[\text{value-side correction}]{\text{Gated Delta Rule-2}}
\bm{v}_t.
\]
Compared with GSA, GSA2 retains the compact two-stage memory structure
and gated memory control while introducing explicit, side-specific
associative correction.

The resulting recurrence remains compatible with efficient parallel training.
We reuse the chunkwise formulation of Gated Delta Rule-2 from GDN2~\citep{hatamizadeh2026gated} and derive a corresponding WY/UT-style formulation for Gated Oja Rule-2.
Both exploit low-rank state transformations so that intra-chunk computation remains dominated by dense matrix multiplications~\citep{bischof1987wy,joffrain2006accumulating}.
GSA2 therefore retains linear-time sequence modeling and constant-size recurrent memory while supporting efficient parallel training.

Empirically, GSA2 performs strongly across language modeling, commonsense reasoning, retrieval, and long-context benchmarks, with particularly clear gains on RULER and LongBench.
Ablations show that pairing key-side Oja correction with value-side Delta correction outperforms correcting either side alone or applying the same correction rule to both memories, with the latter resembling multi-stage fast-weight architectures such as Delta MLPs~\citep{irie2021going} and TRELLIS~\citep{karami2025trellis}.
GSA2 also preserves favorable recurrent length scaling with only a modest constant throughput overhead.

\section{Preliminaries}
\label{sec:preliminaries}

\subsection{Delta-Rule Associative Memory}
\label{sec:prelim_delta}

Additive linear attention continuously superposes new associations in a
fixed-size state, which can cause interference as memory becomes crowded.
DeltaNet~\citep{yang2024parallelizing} instead explicitly corrects the
value associated with the current key:
\begin{equation}
\begin{aligned}
\rmS_t
&=
\rmS_{t-1}
+
\beta_t
\bigl(\vv_t-\rmS_{t-1}\vk_t\bigr)\vk_t^\top
\\
&=
\rmS_{t-1}
\bigl(\rmI-\beta_t\vk_t\vk_t^\top\bigr)
+
\beta_t\vv_t\vk_t^\top .
\end{aligned}
\label{eq:delta_rule}
\end{equation}
where $\beta_t$ controls the update strength and
$\vv_t-\rmS_{t-1}\vk_t$ is the value-prediction error under key
$\vk_t$.
Equivalently, Eq.~\ref{eq:delta_rule} is an online gradient step on
\begin{equation}
\mathcal{L}^{\Delta}_t(\rmS)
=
\frac{1}{2}
\left|
\vv_t-\rmS\vk_t
\right|_2^2 .
\label{eq:delta_objective}
\end{equation}
Thus, the delta rule uses the key to address memory and corrects its
associated value.

Subsequent variants improve this update through adaptive forgetting and
finer-grained memory control.
In particular, GDN2~\citep{hatamizadeh2026gated} decouples erasing old
content from writing new content.
Under the state orientation used in this paper, its channel-wise form is
\begin{equation}
\rmS_t
=
\rmS_{t-1}
\operatorname{Diag}(\alpha_t)
\left(
\rmI-
(\vb_t\odot\vk_t)\vk_t^\top
\right)
+
(\vc_t\odot\vv_t)\vk_t^\top ,
\label{eq:gdn2_preliminary}
\end{equation}
where $\vb_t$ controls erasing along the key-side axis and $\vc_t$
controls writing along the value-side axis.
We reserve $\vw_t$ for the latent-slot representation introduced later.

\subsection{Gated Slot Attention as Two-Stage Associative Memory}
\label{sec:prelim_gsa}

GSA~\citep{zhang2024gated} factorizes a
key--value association through $M$ latent slots.
Using our state orientation,
$\rmS_t^{(1)}\in\mathbb{R}^{M\times D_k}$ stores key--slot associations,
while
$\rmS_t^{(2)}\in\mathbb{R}^{D_v\times M}$ stores slot--value
associations.
With slot-wise forgetting gate
$\alpha_t\in(0,1)^M$, GSA updates the two memories as
\begin{equation}
\rmS_t^{(1)}
=
\operatorname{Diag}(\alpha_t)
\rmS_{t-1}^{(1)}
+
(\mathbf{1}-\alpha_t)\vk_t^\top ,
\label{eq:gsa_stage1}
\end{equation}

\begin{equation}
\rmS_t^{(2)}
=
\rmS_{t-1}^{(2)}
\operatorname{Diag}(\alpha_t)
+
\vv_t(\mathbf{1}-\alpha_t)^\top .
\label{eq:gsa_stage2}
\end{equation}

A query is first mapped to the latent-slot space and then decoded through
the value memory:
\begin{equation}
\vo_t
=
\rmS_t^{(2)}
\operatorname{softmax}
\left(
\rmS_t^{(1)}\vq_t
\right).
\label{eq:gsa_readout}
\end{equation}

This factorization provides an explicit intermediate slot space between
keys and values while retaining compact recurrent memory.
Unlike delta-rule models, however, GSA updates both memories through
gated accumulation rather than explicit error correction.
This distinction motivates the complementary key- and value-side
corrections developed next.

\section{Gated Slot Attention-2}
\label{sec:gsa2}

GSA2 combines a complementary key-side Gated Oja Rule-2 correction with the
value-side Gated Delta Rule-2 correction within the two-stage slot memory of GSA.
We first derive the key-side update and then instantiate the two
corrections through a shared latent-slot representation.

\subsection{Gated Oja Rule: Key-Side Correction}
\label{sec:gated_oja}

The delta rule in Eq.~\ref{eq:delta_rule} corrects the value predicted
from a key.
We consider the complementary problem: correcting the key predicted
from a value.
Specifically, we optimize
\begin{align}
    \mathcal{L}^{K}_t(\bm{S})
    =
    \frac{1}{2}
    \left\|
        \bm{k}_t-\bm{S}^{\top}\bm{v}_t
    \right\|_2^2 .
    \label{eq:oja_objective}
\end{align}
One gradient step gives
\begin{align}
    \bm{S}_t
    &=
    \bm{S}_{t-1}
    +
    \beta_t
    \bm{v}_t
    \left(
        \bm{k}_t-\bm{S}_{t-1}^{\top}\bm{v}_t
    \right)^\top
    \nonumber\\
    &=
    \left(
        \bm{I}-\beta_t\bm{v}_t\bm{v}_t^\top
    \right)
    \bm{S}_{t-1}
    +
    \beta_t\bm{v}_t\bm{k}_t^\top .
    \label{eq:oja_correction}
\end{align}
This left-acting correction has the algebraic form of Oja's normalized
Hebbian update~\citep{oja1982simplified}; we therefore refer to
Eq.~\ref{eq:oja_correction} as the \emph{Oja correction}.
In contrast to the delta rule, which uses a key to correct its
associated value, the Oja correction uses a value to correct its
associated key.

\paragraph{Gated Oja Rule.}
To support adaptive forgetting, we introduce channel-wise decay
$\bm{\alpha}_t$:
\begin{align}
    \bm{S}_t
    =
    \left(
        \bm{I}
        -
        \beta_t\bm{v}_t\bm{v}_t^\top
    \right)
    \operatorname{Diag}(\bm{\alpha}_t)
    \bm{S}_{t-1}
    +
    \beta_t\bm{v}_t\bm{k}_t^\top .
    \label{eq:gated_oja}
\end{align}
We refer to this recurrence as \textbf{OJA Rule}.
\newcommand{\gate}[1]{{\color{blue}#1}}

\begin{table*}[t]
\centering
\caption{
Comparison of recurrent memory mechanisms under the state orientation used in this paper.
\gate{Blue} terms highlight data-dependent memory control.
Mamba-3S and Mamba-3M denote the SISO and MIMO variants of Mamba-3.
Delta-rule updates correct values addressed by keys, whereas OJA and OJA2 apply the complementary value-to-key correction.
GSA and GSA2 factorize memory into key--slot and slot--value states; GSA2 combines Gated OJA Rule-2 and Gated Delta Rule-2 through the shared latent-slot representation $\bm{w}_t$.
}

\label{tab:memory_mechanisms_grouped}

\fontsize{6.6pt}{8.2pt}\selectfont

\setlength{\tabcolsep}{2.2pt}
\renewcommand{\arraystretch}{1.28}

\begin{adjustbox}{max width=\textwidth}
\begin{tabular}{
@{}
>{\centering\arraybackslash}m{0.085\textwidth}
>{\centering\arraybackslash}m{0.160\textwidth}
>{\centering\arraybackslash}m{0.545\textwidth}
>{\centering\arraybackslash}m{0.175\textwidth}
@{}
}

\toprule
\textbf{Model}
&
\textbf{State capacity}
&
\textbf{Gated memory update}
&
\textbf{Output}
\\

\midrule
\multicolumn{4}{@{}l}{\textit{Decay-Gated Additive Updates}} \\
\midrule

Mamba-2
&
$\bm{S}_t\in\mathbb{R}^{D_v\times D_k}$
&
$\bm{S}_t
=
\gate{\alpha_t}\bm{S}_{t-1}
+
\bm{v}_t\bm{k}_t^\top$
&
$\bm{S}_t\bm{q}_t$
\\

Mamba-3S
&
$\bm{S}_t\in\mathbb{R}^{D_v\times D_k}$
&
$\bm{S}_t
=
\gate{\alpha_t}\bm{S}_{t-1}
+
\gate{\eta_t}\bm{v}_{t-1}
\widetilde{\bm{k}}_{t-1}^{\top}
+
\gate{\zeta_t}\bm{v}_{t}
\widetilde{\bm{k}}_{t}^{\top}$
&
$\bm{S}_t\bm{q}_t$
\\

Mamba-3M
&
$\bm{S}_t\in\mathbb{R}^{D_v\times D_k}$
&
$\bm{S}_t
=
\gate{\alpha_t}\bm{S}_{t-1}
+
\sum_{r=1}^{R}
\!\left(
\gate{\eta_{t,r}}
\bm{v}_{t-1}^{(r)}
\widetilde{\bm{k}}_{t-1}^{(r)\top}
+
\gate{\zeta_{t,r}}
\bm{v}_{t}^{(r)}
\widetilde{\bm{k}}_{t}^{(r)\top}
\right)$
&
$\bm{S}_t\bm{q}_t$
\\

\midrule
\multicolumn{4}{@{}l}{\textit{Delta-Rule Error-Correcting Updates}} \\
\midrule

DeltaNet
&
$\bm{S}_t\in\mathbb{R}^{D_v\times D_k}$
&
$\bm{S}_t
=
\bm{S}_{t-1}
\left(
\bm{I}
-
\gate{\beta_t}
\bm{k}_t\bm{k}_t^\top
\right)
+
\gate{\beta_t}
\bm{v}_t\bm{k}_t^\top$
&
$\bm{S}_t\bm{q}_t$
\\

GDN
&
$\bm{S}_t\in\mathbb{R}^{D_v\times D_k}$
&
$\bm{S}_t
=
\bm{S}_{t-1}
\gate{
\alpha_t
\left(
\bm{I}
-
\beta_t
\bm{k}_t\bm{k}_t^\top
\right)
}
+
\gate{\beta_t}
\bm{v}_t\bm{k}_t^\top$
&
$\bm{S}_t\bm{q}_t$
\\

KDA
&
$\bm{S}_t\in\mathbb{R}^{D_v\times D_k}$
&
$\bm{S}_t
=
\bm{S}_{t-1}
\gate{
\operatorname{Diag}(\bm{\alpha}_t)
\left(
\bm{I}
-
\beta_t
\bm{k}_t\bm{k}_t^\top
\right)
}
+
\gate{\beta_t}
\bm{v}_t\bm{k}_t^\top$
&
$\bm{S}_t\bm{q}_t$
\\

GDN2
&
$\bm{S}_t\in\mathbb{R}^{D_v\times D_k}$
&
$\bm{S}_t
=
\bm{S}_{t-1}
\gate{
\operatorname{Diag}(\bm{\alpha}_t)
\left(
\bm{I}
-
(\bm{b}_t\odot\bm{k}_t)
\bm{k}_t^\top
\right)
}
+
(\gate{\bm{c}_t}\odot\bm{v}_t)
\bm{k}_t^\top$
&
$\bm{S}_t\bm{q}_t$
\\

\midrule
\multicolumn{4}{@{}l}{\textit{Oja-Rule Error-Correcting Updates}} \\
\midrule

\textbf{OJA}
&
$\bm{S}_t\in\mathbb{R}^{D_v\times D_k}$
&
$\bm{S}_t
=
\gate{
\left(
\bm{I}
-
\beta_t
\bm{v}_t\bm{v}_t^\top
\right)
\operatorname{Diag}(\bm{\alpha}_t)
}
\bm{S}_{t-1}
+
\gate{\beta_t}
\bm{v}_t\bm{k}_t^\top$
&
$\bm{S}_t\bm{q}_t$
\\

\textbf{OJA2}
&
$\bm{S}_t\in\mathbb{R}^{D_v\times D_k}$
&
$\bm{S}_t
=
\gate{
\left(
\bm{I}
-
\bm{v}_t
(\bm{b}_t\odot\bm{v}_t)^\top
\right)
\operatorname{Diag}(\bm{\alpha}_t)
}
\bm{S}_{t-1}
+
\bm{v}_t
(\gate{\bm{c}_t}\odot\bm{k}_t)^\top$
&
$\bm{S}_t\bm{q}_t$
\\

\midrule
\multicolumn{4}{@{}l}{\textit{Two-Stage Slot-Based Updates}} \\
\midrule

\multirow{2}{*}{GSA}
&
$\bm{S}^{(1)}_t\in\mathbb{R}^{M\times D_k}$
&
$\bm{S}^{(1)}_t
=
\gate{\operatorname{Diag}(\bm{\alpha}_t)}
\bm{S}^{(1)}_{t-1}
+
\gate{(\mathbf{1}-\bm{\alpha}_t)}
\bm{k}_t^\top$
&
\multirow{2}{0.175\textwidth}{\centering
$\bm{S}^{(2)}_t
\operatorname{softmax}
(\bm{S}^{(1)}_t\bm{q}_t)$
}
\\

&
$\bm{S}^{(2)}_t\in\mathbb{R}^{D_v\times M}$
&
$\bm{S}^{(2)}_t
=
\bm{S}^{(2)}_{t-1}
\gate{\operatorname{Diag}(\bm{\alpha}_t)}
+
\bm{v}_t
\gate{(\mathbf{1}-\bm{\alpha}_t)^\top}$
&
\\

\addlinespace[0.2em]

\multirow{2}{*}{\textbf{GSA2}}
&
$\bm{S}^{(1)}_t\in\mathbb{R}^{M\times D_k}$
&
\mbox{
$\bm{S}^{(1)}_t
=
\gate{
\left(
\bm{I}
-
\bm{w}_t
(\bm{b}^{(1)}_t\odot\bm{w}_t)^\top
\right)
\operatorname{Diag}(\bm{\alpha}^{(1)}_t)
}
\bm{S}^{(1)}_{t-1}
+
\bm{w}_t
(\gate{\bm{c}^{(1)}_t}\odot\bm{k}_t)^\top$
}
&
\multirow{2}{0.175\textwidth}{\centering
$\bm{S}^{(2)}_t
\operatorname{silu}
(\bm{S}^{(1)}_t\bm{q}_t)$
}
\\

&
$\bm{S}^{(2)}_t\in\mathbb{R}^{D_v\times M}$
&
$\bm{S}^{(2)}_t
=
\bm{S}^{(2)}_{t-1}
\gate{
\operatorname{Diag}(\bm{\alpha}^{(2)}_t)
\!\left(
\bm{I}
-
(\bm{b}^{(2)}_t\odot\bm{w}_t)
\bm{w}_t^\top
\right)
}
+
(\gate{\bm{c}^{(2)}_t}\odot\bm{v}_t)
\bm{w}_t^\top$
&
\\

\bottomrule
\end{tabular}
\end{adjustbox}

\vspace{0.4em}

\end{table*}
\paragraph{Gated Oja Rule-2.}
Following the erase--write decoupling of GDN2, we further replace the
shared scalar $\beta_t$ with an erase gate $\bm{b}_t$ and a write gate
$\bm{c}_t$:
\begin{align}
    \bm{S}_t
    =
    \left(
        \bm{I}
        -
        \bm{v}_t
        (\bm{b}_t\odot\bm{v}_t)^\top
    \right)
    \operatorname{Diag}(\bm{\alpha}_t)
    \bm{S}_{t-1}
    +
    \bm{v}_t
    (\bm{c}_t\odot\bm{k}_t)^\top .
    \label{eq:oja2}
\end{align}
We call this variant \textbf{OJA2}.
Here, $\bm{b}_t$ controls erasing on the value-side axis, while
$\bm{c}_t$ selects the key channels written into memory.
When
$\bm{b}_t=\beta_t\mathbf{1}_{D_v}$ and
$\bm{c}_t=\beta_t\mathbf{1}_{D_k}$,
where $\mathbf{1}_{D_k}$ denotes the all-ones vector in $\mathbb{R}^{D_k}$,
Eq.~\ref{eq:oja2} reduces to Eq.~\ref{eq:gated_oja}.

Together, OJA2 and GDN2 provide complementary error-correcting
updates: OJA2 corrects the key associated with a value, whereas GDN2
corrects the value associated with a key.

\subsection{Two-Stage Memory with Shared Slots}
\label{sec:gsa2_formulation}

We combine the complementary OJA-2 and Gated Delta Rule-2 updates
within the two-stage memory structure of GSA.
Let
$\bm{S}^{(1)}_t\in\mathbb{R}^{M\times D_k}$ and
$\bm{S}^{(2)}_t\in\mathbb{R}^{D_v\times M}$,
and let $\bm{w}_t\in\mathbb{R}^{M}$ denote a shared latent-slot representation.
It serves as the value-side vector in the first memory and as the
addressing vector in the second, thereby coupling the two stages.
The first-stage update is
\begin{align}
    \bm{S}^{(1)}_t
    &=
    \left(
        \bm{I}
        -
        \bm{w}_t
        (\bm{b}^{(1)}_t\odot\bm{w}_t)^\top
    \right)
    \operatorname{Diag}(\bm{\alpha}^{(1)}_t)
    \bm{S}^{(1)}_{t-1}
    +
    \bm{w}_t
    (\bm{c}^{(1)}_t\odot\bm{k}_t)^\top .
    \label{eq:gsa2_stage1}
\end{align}
It therefore performs key-side correction between the latent slot and
the input key.

The second stage uses the same $\bm{w}_t$ as the addressing key of a
GDN2 memory:
\begin{align}
    \bm{S}^{(2)}_t
    &=
    \bm{S}^{(2)}_{t-1}
    \operatorname{Diag}(\bm{\alpha}^{(2)}_t)
    \left(
        \bm{I}
        -
        (\bm{b}^{(2)}_t\odot\bm{w}_t)
        \bm{w}_t^\top
    \right)
    +
    (\bm{c}^{(2)}_t\odot\bm{v}_t)
    \bm{w}_t^\top .
    \label{eq:gsa2_stage2}
\end{align}
It therefore performs the complementary value-side correction between
the shared slot and the value.
Together, Eqs.~\ref{eq:gsa2_stage1} and~\ref{eq:gsa2_stage2}
instantiate the two-sided associative correction introduced in
Section~\ref{sec_introduction}, with the shared latent slot coupling
key-side and value-side memory editing.

The two memories use independent control gates:
$\bm{b}^{(1)}_t,\bm{b}^{(2)}_t\in[0,1]^M$ are slot-wise erase gates,
while
$\bm{c}^{(1)}_t\in[0,1]^{D_k}$ and
$\bm{c}^{(2)}_t\in[0,1]^{D_v}$
are key- and value-channel write gates.
Thus, GSA2 extends decoupled erase--write control from a single
associative state to both stages of a slot memory.

Finally, GSA2 reads the two memories sequentially:
\begin{align}
    \bm{h}_t
    &=
    \operatorname{silu}
    \left(
        \bm{S}^{(1)}_t\bm{q}_t
    \right),
    &
    \bm{o}_t
    &=
    \bm{S}^{(2)}_t\bm{h}_t .
    \label{eq:gsa2_readout}
\end{align}
Unlike GSA, which applies softmax over the intermediate slot
activations, GSA2 uses SiLU so that corrected slots can contribute
independently rather than compete under a normalized simplex.

Table~\ref{tab:memory_mechanisms_grouped} summarizes the recurrent
memory mechanisms considered in this work under a unified state
orientation.
The comparison highlights the progression from decay-gated additive
updates, to Delta- and Oja-based error-correcting updates, and finally
to the two-stage slot-based memory of GSA2.
In particular, GSA2 combines Gated Oja Rule-2 and Gated Delta Rule-2 through a
shared latent-slot representation, yielding explicit correction on both
sides of the association.

\begin{figure*}[t]
    \centering
    \includegraphics[width=\textwidth]{figures/gated_slots_attention_2.pdf}
    \caption{
Architecture of \textbf{GSA2}.
Left: the hybrid backbone interleaves GSA2, MLP, and SWA blocks.
Right: the GSA2 token mixer.
The query and key branches apply linear projection, short convolution, SiLU, and L2 normalization, while the value branch follows the same pipeline without L2 normalization.
The shared-slot branch uses a low-rank two-layer projection followed by short convolution, SiLU, and L2 normalization.
Gated Oja Rule-2 and Gated Delta Rule-2 perform key- and value-side correction through shared latent slots, with separate decay, erase, and write gates controlling the two recurrent memories.
The output is RMS-normalized, modulated by a sigmoid gate from a low-rank two-layer projection, and linearly projected.
}
    \label{fig:gsa2_model}
\end{figure*}
\subsection{Hardware-Efficient Chunkwise Training}
\label{sec:gsa2_parallel}

GSA2 couples Gated Oja Rule-2 with Gated Delta Rule-2, and efficient
training therefore requires chunkwise forms for both recurrent stages.
For the Gated Delta Rule-2 stage, we directly adopt the
hardware-efficient implementation of GDN2~\citep{hatamizadeh2026gated}.
We therefore focus below on deriving the corresponding chunkwise form
for Gated Oja Rule and Gated Oja Rule-2.

The recurrence is sequential tokenwise, with diagonal decay followed by rank-one state transformations. Within each chunk, the accumulated rank-one transitions have the form
\begin{align}
    \prod_{i=1}^{r}
    \left(
        \bm{I}-\bm{a}_i\bm{z}_i^\top
    \right)
    =
    \bm{I}-\bm{Y}_r\bm{Z}_r^\top ,
    \label{eq:gsa2_wy}
\end{align}
where the right-hand side is a compact WY representation
~\citep{bischof1987wy}.
The channel-wise decay terms can be absorbed into the corresponding
rank-one factors, while the auxiliary matrices can be computed using
triangular solves.
\begin{table*}[t]
    \centering
    
    \caption{
    Performance on language modeling and zero-shot commonsense reasoning.
All non-Transformer hybrid models use SWA.
$^\ast$ Recurrent baseline results are taken from GDN2~\citep{hatamizadeh2026gated}; OJA, OJA2, GSA, and GSA2 are evaluated in this work.}

    \begin{adjustbox}{width=1.0\textwidth,center}
    \begin{small}

    \renewcommand{\arraystretch}{1.35}
    \renewcommand{\multirowsetup}{\centering}
    \setlength{\tabcolsep}{3.2pt}

    \begin{tabular}{
        l|
        cc|
        ccccccccc>{\columncolor{red!7}}c
    }

        \toprule

        \multirow{3}{*}{\textbf{Model}}
        & \multicolumn{2}{c|}{\textbf{Language Modeling}}
        & \multicolumn{10}{c}{\textbf{Zero-shot Commonsense Reasoning}}
        \\

        \cmidrule(lr){2-3}
        \cmidrule(lr){4-13}

        &
        \textbf{Wiki.}
        & \textbf{LMB.}
        & \textbf{LMB.}
        & \textbf{PIQA}
        & \textbf{Hella.}
        & \textbf{Wino.}
        & \textbf{ARC-e}
        & \textbf{ARC-c}
        & \textbf{OBQA}
        & \textbf{SIQA}
        & \textbf{BoolQ}
        & \textbf{Avg.}
        \\

        &
        ppl $\downarrow$
        & ppl $\downarrow$
        & acc $\uparrow$
        & acc $\uparrow$
        & acc$_{\mathrm{n}}$ $\uparrow$
        & acc $\uparrow$
        & acc $\uparrow$
        & acc $\uparrow$
        & acc $\uparrow$
        & acc $\uparrow$
        & acc $\uparrow$
        & acc $\uparrow$
        \\

        \midrule
        \midrule

        \multicolumn{13}{l}{\textit{Recurrent models}} \\

        \hspace{2mm} Mamba-2$^\ast$
        & 16.79 & 12.38
        & 45.24 & 72.58 & 55.51 & 55.33
        & 70.68 & 35.26 & \underline{31.00}
        & 40.63 & 60.19 & 51.82
        \\

        \hspace{2mm} Gated DeltaNet$^\ast$
        & 16.40 & 11.89
        & \underline{49.62} & 72.31 & 56.50 & 56.75
        & 68.81 & 35.15 & 30.20
        & 40.53 & 58.78 & 52.07
        \\

        \hspace{2mm} KDA$^\ast$
        & 16.81 & 11.68
        & 48.13 & 72.09 & 55.75 & 55.72
        & 70.83 & 35.92 & 30.40
        & 40.99 & \underline{60.67} & 52.28
        \\

        \hspace{2mm} Mamba-3 (SISO)$^\ast$
        & 16.24 & 11.84
        & 47.84 & 73.01 & 59.39 & 57.46
        & 72.73 & 37.37 & 27.20
        & \textbf{41.97} & \textbf{60.80} & 53.08
        \\

        \hspace{2mm} Mamba-3 (MIMO)$^\ast$
        & 16.17 & \underline{11.10}
        & 49.21 & \textbf{75.08} & \underline{59.51} & 57.85
        & \textbf{74.12} & \textbf{39.16} & 28.80
        & \underline{41.81} & 54.25 & \underline{53.31}
        \\

        \hspace{2mm} Gated DeltaNet-2$^\ast$
        & \underline{15.90} & 11.41
        & 48.09 & 72.80 & 56.84 & 57.85
        & 72.43 & 38.23 & \textbf{31.60}
        & 40.58 & 59.54 & 53.11
        \\

        \hspace{2mm} OJA
        & 18.14 & 14.47
        & 44.05 & 71.16 & 54.48 & 56.27
        & 68.90 & 33.87 & 27.60
        & \textbf{41.97} & 58.10 & 50.71
        \\

        \hspace{2mm} OJA2
        & 18.28 & 14.84
        & 43.95 & 71.49 & 54.18 & 54.70
        & 68.69 & 33.79 & 27.00
        & 40.17 & 57.37 & 50.15
        \\

        \hspace{2mm} Gated Slot Attention
        & 16.52 & 11.22
        & 49.49 & \underline{74.05} & \textbf{60.22}
        & \underline{59.19}
        & \underline{73.61} & 37.80 & 30.40
        & 39.46 & 55.35 & 53.28
        \\

        \rowcolor{blue!10}
        \hspace{2mm} Gated Slot Attention-2
        & \textbf{15.69} & \textbf{10.68}
        & \textbf{50.07} & 73.56 & 59.28
        & \textbf{60.77}
        & 72.77 & \underline{38.48} & 30.00
        & 41.30 & 60.46 & \textbf{54.08}
        \\

        \midrule

        \multicolumn{13}{l}{\textit{Attention or hybrid models}} \\

        \rowcolor{gray!30}
        \hspace{2mm} Transformer
        & 16.60 & 12.56
        & 49.08 & 72.09 & 57.40 & 56.67
        & 69.95 & 35.07 & 27.20
        & 39.87 & 59.36 & 51.85
        \\

        \hspace{2mm} Mamba-2
        & \textbf{15.85} & 13.62
        & 47.37 & \textbf{73.78} & 58.81 & 58.48
        & 70.83 & \underline{37.37} & 29.20
        & 41.15 & 49.39 & 51.82
        \\

        \hspace{2mm} Gated DeltaNet
        & 16.10 & 12.02
        & 48.69 & 71.98 & 57.46 & 58.96
        & 71.80 & 35.75 & 29.20
        & \underline{41.61} & 57.83 & 52.59
        \\

        \hspace{2mm} KDA
        & 16.22 & 11.49
        & 49.39 & 73.23 & 58.95 & 57.38
        & 71.93 & 37.20 & 29.40
        & 41.10 & 57.58 & 52.91
        \\

        \hspace{2mm} Mamba-3 (SISO)
        & 16.00 & 11.57
        & 49.27 & \underline{73.34} & \underline{59.01}
        & 58.88
        & 71.17 & 36.18 & 28.40
        & \underline{41.61} & 58.26 & 52.90
        \\

        \hspace{2mm} Mamba-3 (MIMO)
        & \underline{15.94} & 11.75
        & 49.51 & 72.80 & \textbf{59.45}
        & 57.93
        & \textbf{72.90} & \textbf{37.63} & \textbf{30.40}
        & \textbf{41.97} & 60.12 & \underline{53.63}
        \\

        \hspace{2mm} Gated DeltaNet-2
        & 16.26 & 11.93
        & 49.49 & 73.01 & 58.09 & 57.46
        & 70.83 & 35.67 & 29.40
        & 39.97 & 59.94 & 52.65
        \\

        \hspace{2mm} OJA
        & 16.53 & 12.12
        & 48.92 & 72.03 & 58.45 & 56.67
        & 70.79 & 34.98 & 28.60
        & 38.08 & \underline{60.86} & 52.15
        \\

        \hspace{2mm} OJA2
        & 16.33 & 12.47
        & 48.83 & 72.47 & 57.78 & 56.35
        & 69.95 & 37.29 & 26.40
        & 38.84 & 59.88 & 51.98
        \\

        \hspace{2mm} Gated Slot Attention
        & 15.98 & \underline{11.36}
        & \underline{49.52} & 72.85 & 58.76
        & \underline{59.35}
        & \underline{71.97} & 36.01 & 28.20
        & 38.28 & 59.05 & 52.67
        \\

        \rowcolor{blue!10}
        \hspace{2mm} Gated Slot Attention-2
        & 17.23 & \textbf{10.80}
        & \textbf{51.76} & 72.91 & 58.92
        & \textbf{59.75}
        & 71.51 & 37.03 & \underline{30.20}
        & 40.99 & \textbf{62.54} & \textbf{53.96}
        \\

        \bottomrule

    \end{tabular}

    \end{small}
    \end{adjustbox}

    \label{tab:lm_commonsense}

\end{table*}
For Gated Oja Rule-2, the resulting chunk-boundary update can be written as
\begin{align}
    \bm{S}_{[t+1]}
    &=
    \operatorname{Diag}
    \left(
        \bm{\alpha}_{[t]}^{1:C}
    \right)
    \bm{S}_{[t]}
    +
    \left(
        \bm{A}_{[t]}^{i+1\rightarrow C}
        \odot
        \bm{V}_{[t]}
    \right)^\top
    \left(
        \bm{U}_{[t]}
        -
        \bm{W}_{[t]}\bm{S}_{[t]}
    \right),
    \label{eq:gsa2_chunk_state}
\end{align}
where $\bm{A}_{[t]}^{i+1\rightarrow C}$ collects the corresponding
suffix-decay factors, and $\bm{U}_{[t]}$ and $\bm{W}_{[t]}$ are
chunk-level auxiliary matrices obtained through the UT transform~\citep{joffrain2006accumulating}.
The uppercase $\bm{W}_{[t]}$ denotes this auxiliary matrix and should
not be confused with the token-level shared latent-slot representation
$\bm{w}_t$ in GSA2.
The corresponding intra-chunk outputs admit an analogous matrix form,
so computation within each chunk is dominated by dense matrix
multiplications.
Detailed WY/UT derivations of Gated Oja Rule and Gated Oja Rule-2 are given in
Appendices~\ref{app:oja} and \ref{app:oja2};
Eq.~\ref{eq:oja-matrix} gives the base chunkwise form, and
Eq.~\ref{eq:oja2-ut} the Gated Oja Rule-2 auxiliary factors.

\subsection{Block Design and Model Families}
\label{subsec:block_design}

\paragraph{GSA2 token mixer.}
GSA2 is used as the recurrent token mixer in a standard
Transformer-style residual block. Fig.~\ref{fig:gsa2_model} (right)
shows its block design.
For the two-stage recurrent update, $\{\bm{q}_t,\bm{k}_t,\bm{v}_t\}$
are produced by linear projection, short causal convolution, and SiLU,
with L2 normalization applied to $\bm{q}_t$ and $\bm{k}_t$ for stability.
The shared latent-slot representation $\bm{w}_t$ is produced by a
low-rank projection followed by short causal convolution, SiLU, and L2
normalization.
Separate branches produce the stage-specific decay, erase, and write
gates
$\{\bm{\alpha}^{(i)}_t,\bm{b}^{(i)}_t,\bm{c}^{(i)}_t\}_{i=1}^{2}$.
The recurrent output is RMS-normalized, modulated by a sigmoid output
gate, and projected back to the model dimension.

\paragraph{Recurrent and hybrid models.}
We evaluate both recurrent and hybrid GSA2 architectures.
The recurrent model alternates GSA2 token mixers with MLP layers under
the standard residual backbone.
The hybrid model follows the same layer pattern as Gated DeltaNet-2,
interleaving GSA2 with Sliding-Window Attention (SWA), as shown in Fig.~\ref{fig:gsa2_model} (left).
A repeated hybrid cell consists of a GSA2 token mixer, an MLP layer,
an SWA layer, and another MLP layer.
GSA2 compresses long-range information into a constant-size recurrent
state, while SWA provides exact interactions within a bounded local
window.
With a fixed window size, the hybrid architecture preserves linear
scaling with sequence length and maintains a bounded attention cache.

\section{Experiments}
\label{sec_exp}
\paragraph{Setup.}
We evaluate each recurrent mixer in both recurrent-only and hybrid
settings, with the latter interleaving the recurrent mixer with
SWA as described in
Section~\ref{subsec:block_design}.
We compare against Transformer, Mamba-2, GDN, KDA,
Mamba-3 (SISO and MIMO), GDN2, GSA, OJA, and OJA2. Models trained in this work contain approximately 1.3B parameters
and are trained on 100B FineWeb-Edu tokens~\citep{penedo2024fineweb}.
Following GDN2~\citep{hatamizadeh2026gated}, we use AdamW with a
peak learning rate of $4\times10^{-4}$, weight decay of 0.1,
gradient clipping at 1.0, cosine decay, a 1B-token warm-up, and a
global batch size of 0.5M tokens.
The training length is 4K tokens, and hybrid models use a 2K SWA
window.
Additional details are provided in the Appendix~\ref{app:exp}.

\paragraph{Language modeling and common-sense reasoning.}
Table~\ref{tab:lm_commonsense} reports WikiText and LAMBADA perplexity
\citep{merity2016pointer,paperno2016lambada}, zero-shot LAMBADA
accuracy, and the common-sense suite from PIQA through BoolQ
\citep{bisk2020piqa,zellers2019hellaswag,sakaguchi2021winogrande,
clark2018think,mihaylov2018can,sap2019social,clark2019boolq}.
GSA2 achieves the best common-sense average in both recurrent and
hybrid settings while remaining strong on language modeling.
The gains are consistent against strong recurrent baselines such as
Gated DeltaNet-2, OJA2, and Mamba-3, and persist when combined with SWA.

\begin{wraptable}{r}{0.50\textwidth}
    \centering
    \caption{
        Real-world retrieval at 2K context. All non-Transformer hybrid models use SWA.
    }
    \captionsetup{
        font=scriptsize,
        justification=raggedright,
        singlelinecheck=false,
        skip=3pt
    }

    \scriptsize
    \renewcommand{\arraystretch}{1.15}
    \setlength{\tabcolsep}{1.2pt}

    \resizebox{\linewidth}{!}{%
    \begin{tabular}{
        @{}l|
        cccccc>{\columncolor{red!7}}c
        @{}
    }

        \toprule

        \multirow{2}{*}{\textbf{Model}}
        & \multicolumn{7}{c}{\textbf{Real-world Retrieval (2K)}} \\

        \cmidrule(lr){2-8}

        &
        \textbf{SWDE}
        & \textbf{SQD}
        & \textbf{FDA}
        & \textbf{TQA}
        & \textbf{NQ}
        & \textbf{DROP}
        & \textbf{Avg.}
        \\

        &
        acc $\uparrow$
        & acc $\uparrow$
        & acc $\uparrow$
        & acc $\uparrow$
        & acc $\uparrow$
        & acc $\uparrow$
        & acc $\uparrow$
        \\

        \midrule
        \midrule

        \multicolumn{8}{l}{\textit{Recurrent models}} \\

        \hspace{1mm} Mamba-2$^\ast$
        & 17.24 & 32.38 & 14.53
        & 58.35 & 18.91 & 19.60 & 26.84
        \\

        \hspace{1mm} Gated DeltaNet$^\ast$
        & 17.90 & 32.67 & 18.52
        & 59.60 & \textbf{20.16} & 19.69 & 28.09
        \\

        \hspace{1mm} KDA$^\ast$
        & 22.49 & 35.10 & 14.90
        & 58.12 & 19.58 & \textbf{21.80} & 28.67
        \\

        \hspace{1mm} Mamba-3 (SISO)$^\ast$
        & \underline{27.36} & 34.62 & 10.53
        & 60.37 & 15.93 & \underline{21.75} & 28.43
        \\

        \hspace{1mm} Mamba-3 (MIMO)$^\ast$
        & 26.01 & 34.85 & 16.97
        & 60.60 & 15.52 & 19.65 & 28.93
        \\

        \hspace{1mm} Gated DeltaNet-2$^\ast$
        & 23.65 & \textbf{36.75} & \textbf{19.98}
        & \underline{61.37} & \underline{19.64}
        & 17.87 & \underline{29.88}
        \\

        \hspace{1mm} OJA
        & 22.14 & 33.21 & 7.71
        & 58.77 & 15.05 & 18.83 & 25.95
        \\

        \hspace{1mm} OJA2
        & 20.61 & 33.75 & 12.07
        & 58.18 & 14.22 & 18.59 & 26.24
        \\

        \hspace{1mm} Gated Slot Attention
        & 17.82 & 32.47 & 12.79
        & 61.26 & 15.81 & 20.12 & 26.71
        \\

        \rowcolor{blue!10}
        \hspace{1mm} Gated Slot Attention-2
        & \textbf{29.97} & \underline{36.29} & \underline{19.42}
        & \textbf{63.09} & 17.26 & 21.51 & \textbf{31.26}
        \\

        \midrule

        \multicolumn{8}{l}{\textit{Attention or hybrid models}} \\

        \rowcolor{gray!30}
        \hspace{1mm} Transformer
        & 41.58 & 41.12 & 21.42
        & 63.15 & 16.79 & 22.42 & 34.41
        \\

        \hspace{1mm} Mamba-2
        & 36.00 & 36.33 & 26.13
        & 58.47 & 18.47 & 18.74 & 32.36
        \\

        \hspace{1mm} Gated DeltaNet
        & 43.11 & 40.62 & \underline{29.85}
        & 63.45 & 17.90 & 20.60 & 35.92
        \\

        \hspace{1mm} KDA
        & 42.66 & 41.05 & 27.95
        & 62.56 & 17.90 & 22.28 & 35.73
        \\

        \hspace{1mm} Mamba-3 (SISO)
        & 41.67 & 40.25 & 25.50
        & 62.97 & 18.15 & 22.09 & 35.11
        \\

        \hspace{1mm} Mamba-3 (MIMO)
        & 41.49 & 40.28 & 28.58
        & 63.51 & 17.55 & 22.86 & 35.71
        \\

        \hspace{1mm} Gated DeltaNet-2
        & 44.01 & 40.75 & 27.95
        & 61.49 & 17.58 & 20.65 & 35.41
        \\

        \hspace{1mm} OJA
        & 47.25 & \underline{41.76} & 28.86
        & \textbf{64.69} & \underline{18.63}
        & \textbf{23.77} & \textbf{37.49}
        \\

        \hspace{1mm} OJA2
        & \underline{47.61} & 40.78 & \textbf{30.40}
        & 62.91 & 17.71 & \underline{23.72} & \underline{37.19}
        \\

        \hspace{1mm} Gated Slot Attention
        & 35.55 & 40.35 & 19.78
        & \underline{63.57} & 16.69 & 21.37 & 32.89
        \\

        \rowcolor{blue!10}
        \hspace{1mm} Gated Slot Attention-2
        & \textbf{51.31} & \textbf{42.26} & 25.05
        & 62.56 & \textbf{18.78} & 20.22 & 36.70
        \\

        \bottomrule
        
    \end{tabular}%
    }
\vspace{-2em}
    
    \label{tab:realworld_retrieval}

\end{wraptable}

\paragraph{In-context retrieval on real-world tasks.}
Table~\ref{tab:realworld_retrieval} reports recall-heavy real-world tasks
from~\citep{arora2024just}, spanning extraction, question answering,
and distractor-rich evidence.
GSA2 achieves the best average in the recurrent setting, showing that
two-sided correction transfers beyond synthetic retrieval.
In the hybrid setting, OJA and OJA2 become particularly competitive.
Since these tasks use 2K inputs, matching the SWA window, local attention
can directly access most relevant evidence, reducing the burden on the
recurrent state.

\begin{figure*}[t]
    \centering
    \includegraphics[width=\textwidth]{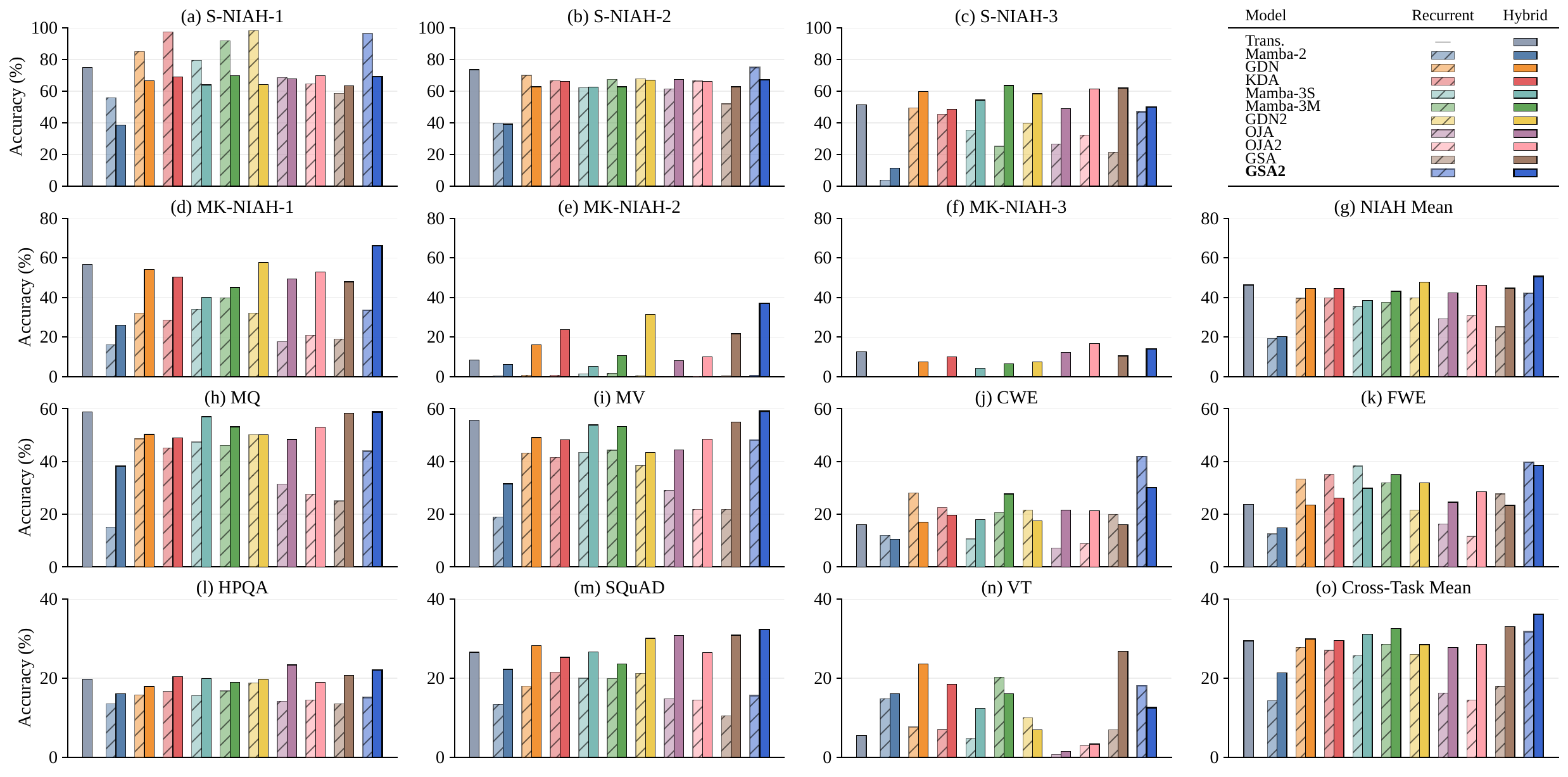}
    \caption{
RULER accuracy of recurrent and attention/hybrid models.
Each task is averaged over 1K--8K contexts.
NIAH Avg. summarizes the 24 S-NIAH and MK-NIAH settings, while Cross-Task Avg.
summarizes the 28 settings from MQ, MV, CWE, FWE, HPQA, SQuAD, and VT.
Colors denote base models; hatched and solid bars indicate recurrent and
attention/hybrid configurations, respectively.
}
\label{fig:ruler_niah}
\end{figure*}

\paragraph{In-context retrieval on synthetic data.}
Fig.~\ref{fig:ruler_niah} reports
RULER results across NIAH and other long-context tasks~\citep{hsieh2024ruler}.
GSA2 achieves the strongest overall retrieval performance in both recurrent
and hybrid settings.
Its gains over GSA, GDN2, and OJA2 show that
two-sided correction is more effective than gated accumulation or
correcting either side alone.
This supports combining key-side Gated Oja Rule-2 correction with value-side
Gated Delta Rule-2 correction in the two-stage memory. Full results are provided in Appendix
Tables~\ref{tab:ruler_niah} and~\ref{tab:ruler_remaining}.

\setlength{\columnsep}{6pt}
\setlength{\intextsep}{3pt}

\begin{table}
    \centering
    \caption{
    LongBench performance across five task categories.
    Avg. is the mean over all 11 tasks.
    All non-Transformer models in the hybrid group use SWA.
    }
    \vspace{0pt}

    \includegraphics[
        width=\linewidth
    ]{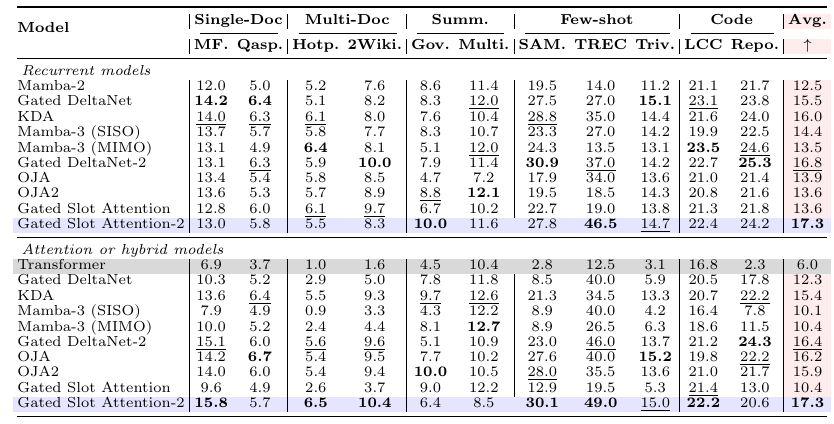}

    \label{tab:longbench_results}

    \vspace{2pt}
\end{table}

\noindent\textbf{Long-context evaluation.}\enspace
Table~\ref{tab:longbench_results} reports
LongBench~\cite{bai2024longbench} results on document understanding,
summarization, few-shot learning, and code.
GSA2 achieves the highest overall average in both recurrent and hybrid
settings.
Together with RULER, these results show that two-sided associative
correction benefits long-context modeling beyond controlled retrieval.

\begin{table*}[t]
\centering
\caption{
Ablation study of GSA2.
Common. averages LAMBADA accuracy and nine zero-shot common-sense reasoning tasks.
S-NIAH, MK-NIAH, and Rem. are averages over the corresponding RULER task groups.
Real-world retrieval averages six tasks at 2K context, and LongBench averages 11 tasks.}

\setlength{\tabcolsep}{1.4pt}
\renewcommand{\arraystretch}{1.2}

\resizebox{\textwidth}{!}{%
\begin{tabular}{cc|cc|c|ccc|c|c}
\toprule

\multicolumn{2}{c|}{\textbf{Ablation configuration}}
& \multicolumn{2}{c|}{\textbf{Language modeling}}
& \textbf{Common.}
& \multicolumn{3}{c|}{\textbf{RULER}}
& \textbf{Real-world retrieval}
& \textbf{LongBench}
\\

\cmidrule(lr){1-2}
\cmidrule(lr){3-4}
\cmidrule(lr){5-5}
\cmidrule(lr){6-8}
\cmidrule(lr){9-9}
\cmidrule(lr){10-10}

\textbf{Stage 1}
& \textbf{Stage 2}
& \textbf{Wiki.}$\downarrow$
& \textbf{LMB.}$\downarrow$
& \textbf{avg}$\uparrow$
& \textbf{S-NIAH}$\uparrow$
& \textbf{MK-NIAH}$\uparrow$
& \textbf{Rem.}$\uparrow$
& \textbf{avg}$\uparrow$
& \textbf{avg}$\uparrow$
\\

\midrule

\multirow{2}{*}{Coupled}
& Coupled
& 16.22
& \underline{10.29}
& 53.25
& 68.2
& 9.9
& 30.0
& 29.60
& \underline{17.1}
\\

& Decoupled
& 16.01
& 10.65
& 52.46
& 68.0
& 10.9
& 24.7
& \underline{31.26}
& 14.8
\\

\midrule

\multirow{2}{*}{Decoupled}
& Coupled
& 16.28
& 11.25
& 53.37
& 65.3
& 9.8
& 28.3
& 30.17
& 16.2
\\

& \textbf{Decoupled}
& \textbf{15.69}
& 10.68
& \textbf{54.08}
& \textbf{72.8}
& \underline{11.4}
& \textbf{31.8}
& \textbf{31.51}
& \textbf{17.3}
\\

\midrule

\multicolumn{2}{c|}{Gated Delta Rule-2 in both updates}
& \underline{15.92}
& \textbf{10.27}
& 53.03
& \underline{72.0}
& \textbf{13.1}
& 29.5
& 29.76
& 16.8
\\

\multicolumn{2}{c|}{Gated Oja Rule-2 in both updates}
& 16.97
& 12.46
& 52.35
& 60.2
& 8.4
& 20.9
& 30.96
& 13.8
\\

\midrule

\multicolumn{2}{c|}{w/o $1/\sqrt{d}$ Scaling}
& 19.76
& 16.98
& 48.50
& 61.8
& 8.3
& 21.8
& 24.73
& 13.0
\\

\multicolumn{2}{c|}{SiLU $\rightarrow$ Softmax}
& 16.92
& 11.36
& 50.66
& 55.6
& 9.7
& 26.1
& 28.14
& 14.0
\\

\midrule

\multicolumn{2}{c|}{Slots $=64$}
& 16.55
& 11.17
& \underline{53.48}
& 62.6
& 8.2
& 30.0
& 29.06
& 16.9
\\

\multicolumn{2}{c|}{Slots $=256$}
& 16.37
& 11.36
& 52.87
& 61.7
& 11.1
& \underline{30.4}
& 30.10
& 15.3
\\

\bottomrule
\end{tabular}%
}

\vspace{1em}

\label{tab:gsa2_ablation}

\end{table*}

\paragraph{Ablation studies.}
Table~\ref{tab:gsa2_ablation} examines the main design choices of GSA2.
All ablation variants use the 1.3B model scale and are trained on 100B tokens.

(1) \emph{Complementary two-sided correction.}
We first compare the correction rules used in the two recurrent memories.
Applying Gated Delta Rule-2 to both memories or Gated Oja Rule-2 to both
memories is less effective overall than GSA2, which combines key-side
Gated Oja Rule-2 with value-side Gated Delta Rule-2 through the shared
latent slots.
\textbf{This supports using complementary correction directions for the
semantically distinct key- and value-side memories rather than repeating
the same correction rule across both.}

(2) \emph{Standalone correction direction.}
When used as standalone recurrent updates, Gated DeltaNet-2 generally
outperforms OJA2, as shown in Tables~\ref{tab:lm_commonsense},
\ref{tab:realworld_retrieval},
\ref{tab:longbench_results},
\ref{tab:ruler_niah}
and \ref{tab:ruler_remaining}.
Thus, the benefit of Gated Oja Rule-2 in GSA2 does not arise from being a
uniformly stronger recurrent update.
Instead, it is most effective as the complementary key-side correction
within the two-stage slot memory, while the Delta rule remains better
suited to value-side correction.

(3) \emph{Erase--write decoupling.}
Decoupling erase and write in both memories gives the strongest overall
performance, while decoupling only one side yields less consistent gains.
This shows that independent erase--write control benefits both key- and
value-side memories.

(4) \emph{Architectural choices.}
Removing the standard $1/\sqrt{d}$ scaling factor or replacing SiLU with
Softmax causes broad degradation.
Varying the number of latent slots further shows that the default
128-slot configuration provides the best overall balance across the
evaluated tasks.

\begin{figure*}[t]
\centering

\definecolor{transformerblue}{RGB}{31,119,180}

\definecolor{mambaorange}{RGB}{230,108,32}
\definecolor{mambaorange2}{RGB}{244,150,72}
\definecolor{mambaorange3}{RGB}{190,78,35}

\definecolor{gdngreen}{RGB}{92,114,50}
\definecolor{kdablue}{RGB}{54,126,177}
\definecolor{gdn2green}{RGB}{50,140,90}

\definecolor{ojapurple}{RGB}{134,96,180}
\definecolor{oja2purple}{RGB}{100,65,150}

\definecolor{gsagray}{RGB}{105,105,105}
\definecolor{gsa2red}{RGB}{200,65,65}

\begin{tikzpicture}
\begin{axis}[
    width=0.99\textwidth,
    height=6.25cm,
    xlabel={Sequence length $\times$ batch size},
    ylabel={Training throughput (Kt/s)},
    xmin=0.9, xmax=4.1,
    ymin=8.0, ymax=15.2,
    xtick={1,2,3,4},
    xticklabels={
        2K$\times$8,
        4K$\times$4,
        8K$\times$2,
        16K$\times$1
    },
    ytick={8,9,10,11,12,13,14,15},
    grid=major,
    major grid style={black!10},
    axis line style={black!60},
    tick style={black!60},
    tick label style={font=\scriptsize},
    label style={font=\small},
    every axis plot/.append style={
        line width=1.35pt,
        mark size=2.4pt
    },
    legend columns=3,
    legend cell align=left,
    legend style={
        at={(0.03,0.03)},
        anchor=south west,
        font=\fontsize{6.5}{7.2}\selectfont,
        draw=black!25,
        fill=white,
        fill opacity=0.96,
        text opacity=1,
        rounded corners=1pt,
        inner xsep=3pt,
        inner ysep=2pt,
        row sep=0.6pt,
        column sep=6pt
    },
    clip=true,
]

% Transformer
\addplot[
    color=transformerblue,
    mark=*,
]
coordinates {
    (1,14.366)
    (2,13.105)
    (3,11.138)
    (4,8.595)
};
\addlegendentry{Transformer}

% Mamba family
\addplot[
    color=mambaorange,
    mark=pentagon*,
]
coordinates {
    (1,14.229)
    (2,13.824)
    (3,13.623)
    (4,13.516)
};
\addlegendentry{Mamba-2}

\addplot[
    color=mambaorange2,
    mark=triangle*,
]
coordinates {
    (1,14.489)
    (2,14.034)
    (3,13.808)
    (4,13.512)
};
\addlegendentry{Mamba-3 SISO}

\addplot[
    color=mambaorange3,
    mark=diamond*,
]
coordinates {
    (1,12.347)
    (2,11.841)
    (3,11.647)
    (4,11.047)
};
\addlegendentry{Mamba-3 MIMO}

% Delta family
\addplot[
    color=gdngreen,
    mark=triangle*,
]
coordinates {
    (1,14.284)
    (2,13.971)
    (3,13.789)
    (4,13.672)
};
\addlegendentry{Gated DeltaNet}

\addplot[
    color=kdablue,
    mark=square*,
]
coordinates {
    (1,13.782)
    (2,13.461)
    (3,13.293)
    (4,13.178)
};
\addlegendentry{KDA}

\addplot[
    color=gdn2green,
    mark=triangle,
]
coordinates {
    (1,13.534)
    (2,13.232)
    (3,13.067)
    (4,12.968)
};
\addlegendentry{Gated DeltaNet-2}

% Oja family
\addplot[
    color=ojapurple,
    mark=o,
]
coordinates {
    (1,13.211)
    (2,12.898)
    (3,12.752)
    (4,12.630)
};
\addlegendentry{OJA}

\addplot[
    color=oja2purple,
    mark=*,
]
coordinates {
    (1,12.932)
    (2,12.638)
    (3,12.494)
    (4,12.376)
};
\addlegendentry{OJA2}

% GSA family
\addplot[
    color=gsagray,
    mark=square,
]
coordinates {
    (1,14.056)
    (2,13.703)
    (3,13.522)
    (4,13.378)
};
\addlegendentry{GSA}

\addplot[
    color=gsa2red,
    mark=star,
    very thick,
    line width=1.8pt,
]
coordinates {
    (1,11.818)
    (2,11.560)
    (3,11.429)
    (4,11.317)
};
\addlegendentry{\textbf{GSA2}}

\end{axis}
\end{tikzpicture}

\caption{
Training throughput of parameter-matched 1.3B models on a single NVIDIA H20 GPU under a fixed 16K-token budget per optimization step.
Sequence length increases from 2K to 16K while batch size decreases correspondingly from 8 to 1.
The throughput evaluation protocol is detailed in
Appendix~\ref{app:training_efficiency}.
}
\label{fig:throughput_gsa2}
\end{figure*}
\paragraph{Throughput comparison.}
Fig.~\ref{fig:throughput_gsa2} reports single-H20 training throughput
for the hybrid 1.30B models under a fixed 16,384-token budget; detailed
experimental settings are given in Appendix~\ref{app:training_efficiency}.
GSA2 preserves the near-flat scaling profile of recurrent mixers as
sequence length grows, dropping only from 11.82 to 11.32 Kt/s
($4.2\%$), while the Transformer degrades sharply from 14.37 to
8.60 Kt/s ($40.2\%$).
OJA, OJA2, and Gated DeltaNet-2 show similarly flat scaling, with
throughput reductions of $4.4\%$, $4.3\%$, and $4.2\%$, respectively.
Relative to OJA2 and Gated DeltaNet-2, GSA2 is approximately $8.6\%$
and $12.6\%$ slower, respectively, reflecting the additional cost of
coupling key-side and value-side error correction through a two-stage
slot memory.
Thus, GSA2 retains the favorable length-scaling behavior of recurrent
attention while paying a moderate constant cost for two-sided
associative memory correction.

\section{Related Work}
\label{sec_related_work}

Efficient sequence models replace quadratic self-attention with recurrent
or linear-time token mixers that maintain fixed-size states.
Representative approaches include S4/S5-style state-space models and
RetNet~\citep{gu2021efficiently,smith2022simplified,orvieto2023resurrecting,sun2023retentive},
Mamba and its successors~\citep{gu2023mamba,dao2024transformers,lahoti2026mamba},
and gated linear RNNs~\citep{yang2023gated,qin2024hgrn2}.
Within this family, Gated Slot Attention (GSA)~\citep{zhang2024gated}
factorizes associative memory through shared latent slots, maintaining
separate key- and value-side memories with gated accumulation. Recurrent states can also be viewed as fast-weight associative memories.
Classical work studied error-correcting alternatives to Hebbian
accumulation~\citep{gardner1988space,prados1989neural}, while modern
models instantiate such correction through the delta rule
~\citep{irie2021going,schlag2021linear,yang2024parallelizing}.
Gated DeltaNet~\citep{yang2025gated} adds adaptive forgetting,
KDA~\citep{team2025kimi} introduces channel-wise decay, and
Gated DeltaNet-2~\citep{hatamizadeh2026gated} decouples erase and write.
In this work, 
GSA2 connects these ideas by introducing complementary Oja- and
Delta-based corrections to the key and value sides of GSA.

\section{Conclusion}
\label{sec_conclusion}

We introduced Gated Slot Attention-2 (GSA2), a two-stage recurrent
attention layer that performs complementary associative correction on
both sides of a shared latent-slot memory.
GSA2 uses Gated Oja Rule-2 to correct key--slot associations in the
first stage and Gated Delta Rule-2 to correct slot--value associations
in the second, while decoupling erase and write control in both stages.
The resulting updates preserve efficient chunkwise training through
compact WY representations while retaining fixed-size recurrent memory.
At the 1.3B scale, GSA2 achieves strong performance across language
modeling, commonsense reasoning, synthetic and real-world retrieval,
and general long-context evaluation, with particularly consistent gains
on RULER and LongBench.
Ablations show that the gains arise from combining complementary
correction directions with decoupled memory editing, rather than simply
repeating the same update rule across both stages.

\subsection*{AI Use Statement}

In this work, we used generative AI tools solely to aid and polish the
writing. We did not use generative AI tools for research design,
methodology, experiments, data analysis, or the generation of results.
All AI-assisted content was reviewed by the authors.
We take full responsibility for the final content of this work.

\subsection*{Ethics statement}

This work studies efficient sequence modeling architectures and does not
involve human subjects or the collection of sensitive personal data.
Experiments are conducted using publicly available training and evaluation
datasets. We are not aware of additional ethical concerns specific to the
proposed method beyond those generally associated with large language models.

\subsection*{Reproducibility statement}
We provide the model architecture, recurrent update rules, and main training settings in the paper. The appendix provides detailed derivations of the proposed recurrent updates, additional training and evaluation details, and the protocol used for the training-throughput measurements. The complete source code will be made publicly available upon acceptance of the paper.



\bibliography{iclr2027_conference}

@article{vaswani2017attention,
  title={Attention is all you need},
  author={Vaswani, Ashish and Shazeer, Noam and Parmar, Niki and Uszkoreit, Jakob and Jones, Llion and Gomez, Aidan N and Kaiser, {\L}ukasz and Polosukhin, Illia},
  journal={Advances in neural information processing systems},
  volume={30},
  year={2017}
}

@inproceedings{katharopoulos2020transformers,
  title={Transformers are rnns: Fast autoregressive transformers with linear attention},
  author={Katharopoulos, Angelos and Vyas, Apoorv and Pappas, Nikolaos and Fleuret, Fran{\c{c}}ois},
  booktitle={International conference on machine learning},
  pages={5156--5165},
  year={2020},
  organization={PMLR}
}

@article{sun2023retentive,
  title={Retentive network: A successor to transformer for large language models},
  author={Sun, Yutao and Dong, Li and Huang, Shaohan and Ma, Shuming and Xia, Yuqing and Xue, Jilong and Wang, Jianyong and Wei, Furu},
  journal={arXiv preprint arXiv:2307.08621},
  year={2023}
}

@article{yang2023gated,
  title={Gated linear attention transformers with hardware-efficient training},
  author={Yang, Songlin and Wang, Bailin and Shen, Yikang and Panda, Rameswar and Kim, Yoon},
  journal={arXiv preprint arXiv:2312.06635},
  year={2023}
}

@article{gu2023mamba,
  title={Mamba: Linear-time sequence modeling with selective state spaces},
  author={Gu, Albert and Dao, Tri},
  journal={arXiv preprint arXiv:2312.00752},
  year={2023}
}

@article{dao2024transformers,
  title={Transformers are SSMs: Generalized models and efficient algorithms through structured state space duality},
  author={Dao, Tri and Gu, Albert},
  journal={arXiv preprint arXiv:2405.21060},
  year={2024}
}

@article{hu2025comba,
  title={Comba: Improving Bilinear RNNs with Closed-loop Control},
  author={Hu, Jiaxi and Pan, Yongqi and Du, Jusen and Lan, Disen and Tang, Xiaqiang and Wen, Qingsong and Liang, Yuxuan and Sun, Weigao},
  journal={arXiv preprint arXiv:2506.02475},
  year={2025}
}

@inproceedings{lahoti2026mamba,
  title={Mamba-3: Improved sequence modeling using state space principles},
  author={Lahoti, Aakash Sunil and Li, Kevin and Chen, Berlin and Wang, Caitlin and Bick, Aviv and Kolter, Zico and Dao, Tri and Gu, Albert},
  booktitle={International Conference on Learning Representations},
  volume={2026},
  pages={86173--86202},
  year={2026}
}

@article{qin2024hgrn2,
  title={Hgrn2: Gated linear rnns with state expansion},
  author={Qin, Zhen and Yang, Songlin and Sun, Weixuan and Shen, Xuyang and Li, Dong and Sun, Weigao and Zhong, Yiran},
  journal={arXiv preprint arXiv:2404.07904},
  year={2024}
}

@article{yang2024parallelizing,
  title={Parallelizing linear transformers with the delta rule over sequence length},
  author={Yang, Songlin and Wang, Bailin and Zhang, Yu and Shen, Yikang and Kim, Yoon},
  journal={Advances in neural information processing systems},
  volume={37},
  pages={115491--115522},
  year={2024}
}

@inproceedings{yang2025gated,
  title={Gated delta networks: Improving mamba2 with delta rule},
  author={Yang, Songlin and Kautz, Jan and Hatamizadeh, Ali},
  booktitle={International Conference on Learning Representations},
  volume={2025},
  pages={29687--29707},
  year={2025}
}

@article{team2025kimi,
  title={Kimi linear: An expressive, efficient attention architecture},
  author={Team, Kimi and Zhang, Yu and Lin, Zongyu and Yao, Xingcheng and Hu, Jiaxi and Meng, Fanqing and Liu, Chengyin and Men, Xin and Yang, Songlin and Li, Zhiyuan and others},
  journal={arXiv preprint arXiv:2510.26692},
  year={2025}
}

@article{hatamizadeh2026gated,
  title={Gated DeltaNet-2: Decoupling erase and write in linear attention},
  author={Hatamizadeh, Ali and Choi, Yejin and Kautz, Jan},
  journal={arXiv preprint arXiv:2605.22791},
  year={2026}
}

@article{zhang2024gated,
  title={Gated slot attention for efficient linear-time sequence modeling},
  author={Zhang, Yu and Yang, Songlin and Zhu, Ruijie and Zhang, Yue and Cui, Leyang and Wang, Yiqiao and Wang, Bolun and Shi, Freda and Wang, Bailin and Bi, Wei and others},
  journal={Advances in Neural Information Processing Systems},
  volume={37},
  pages={116870--116898},
  year={2024}
}

@article{bischof1987wy,
  title={The WY representation for products of Householder matrices},
  author={Bischof, Christian and Van Loan, Charles},
  journal={SIAM Journal on Scientific and Statistical Computing},
  volume={8},
  number={1},
  pages={s2--s13},
  year={1987},
  publisher={SIAM}
}

@article{oja1982simplified,
  title={Simplified neuron model as a principal component analyzer},
  author={Oja, Erkki},
  journal={Journal of mathematical biology},
  volume={15},
  number={3},
  pages={267--273},
  year={1982},
  publisher={Springer}
}

@article{penedo2024fineweb,
  title={The fineweb datasets: Decanting the web for the finest text data at scale},
  author={Penedo, Guilherme and Kydl{\'\i}{\v{c}}ek, Hynek and Lozhkov, Anton and Mitchell, Margaret and Raffel, Colin and Von Werra, Leandro and Wolf, Thomas and others},
  journal={Advances in Neural Information Processing Systems},
  volume={37},
  pages={30811--30849},
  year={2024}
}

@article{merity2016pointer,
  title={Pointer sentinel mixture models},
  author={Merity, Stephen and Xiong, Caiming and Bradbury, James and Socher, Richard},
  journal={arXiv preprint arXiv:1609.07843},
  year={2016}
}

@inproceedings{paperno2016lambada,
  title={The LAMBADA dataset: Word prediction requiring a broad discourse context},
  author={Paperno, Denis and Kruszewski, Germ{\'a}n and Lazaridou, Angeliki and Pham, Ngoc-Quan and Bernardi, Raffaella and Pezzelle, Sandro and Baroni, Marco and Boleda, Gemma and Fern{\'a}ndez, Raquel},
  booktitle={Proceedings of the 54th annual meeting of the association for computational linguistics (volume 1: Long papers)},
  pages={1525--1534},
  year={2016}
}

@inproceedings{bisk2020piqa,
  title={Piqa: Reasoning about physical commonsense in natural language},
  author={Bisk, Yonatan and Zellers, Rowan and Gao, Jianfeng and Choi, Yejin and others},
  booktitle={Proceedings of the AAAI conference on artificial intelligence},
  volume={34},
  number={05},
  pages={7432--7439},
  year={2020}
}

@inproceedings{zellers2019hellaswag,
  title={Hellaswag: Can a machine really finish your sentence?},
  author={Zellers, Rowan and Holtzman, Ari and Bisk, Yonatan and Farhadi, Ali and Choi, Yejin},
  booktitle={Proceedings of the 57th annual meeting of the association for computational linguistics},
  pages={4791--4800},
  year={2019}
}

@article{sakaguchi2021winogrande,
  title={Winogrande: An adversarial winograd schema challenge at scale},
  author={Sakaguchi, Keisuke and Bras, Ronan Le and Bhagavatula, Chandra and Choi, Yejin},
  journal={Communications of the ACM},
  volume={64},
  number={9},
  pages={99--106},
  year={2021},
  publisher={ACM New York, NY, USA}
}

@article{clark2018think,
  title={Think you have solved question answering? try arc, the ai2 reasoning challenge},
  author={Clark, Peter and Cowhey, Isaac and Etzioni, Oren and Khot, Tushar and Sabharwal, Ashish and Schoenick, Carissa and Tafjord, Oyvind},
  journal={arXiv preprint arXiv:1803.05457},
  year={2018}
}

@inproceedings{mihaylov2018can,
  title={Can a suit of armor conduct electricity? a new dataset for open book question answering},
  author={Mihaylov, Todor and Clark, Peter and Khot, Tushar and Sabharwal, Ashish},
  booktitle={Proceedings of the 2018 conference on empirical methods in natural language processing},
  pages={2381--2391},
  year={2018}
}

@inproceedings{sap2019social,
  title={Social IQa: Commonsense reasoning about social interactions},
  author={Sap, Maarten and Rashkin, Hannah and Chen, Derek and Le Bras, Ronan and Choi, Yejin},
  booktitle={Proceedings of the 2019 conference on empirical methods in natural language processing and the 9th international joint conference on natural language processing (EMNLP-IJCNLP)},
  pages={4463--4473},
  year={2019}
}

@inproceedings{clark2019boolq,
  title={Boolq: Exploring the surprising difficulty of natural yes/no questions},
  author={Clark, Christopher and Lee, Kenton and Chang, Ming-Wei and Kwiatkowski, Tom and Collins, Michael and Toutanova, Kristina},
  booktitle={Proceedings of the 2019 Conference of the North American Chapter of the Association for Computational Linguistics: Human Language Technologies, Volume 1 (Long and Short Papers)},
  pages={2924--2936},
  year={2019}
}

@article{hsieh2024ruler,
  title={RULER: What's the real context size of your long-context language models?},
  author={Hsieh, Cheng-Ping and Sun, Simeng and Kriman, Samuel and Acharya, Shantanu and Rekesh, Dima and Jia, Fei and Zhang, Yang and Ginsburg, Boris},
  journal={arXiv preprint arXiv:2404.06654},
  year={2024}
}

@article{arora2024just,
  title={Just read twice: closing the recall gap for recurrent language models},
  author={Arora, Simran and Timalsina, Aman and Singhal, Aaryan and Spector, Benjamin and Eyuboglu, Sabri and Zhao, Xinyi and Rao, Ashish and Rudra, Atri and R{\'e}, Christopher},
  journal={arXiv preprint arXiv:2407.05483},
  year={2024}
}

@inproceedings{bai2024longbench,
  title={Longbench: A bilingual, multitask benchmark for long context understanding},
  author={Bai, Yushi and Lv, Xin and Zhang, Jiajie and Lyu, Hongchang and Tang, Jiankai and Huang, Zhidian and Du, Zhengxiao and Liu, Xiao and Zeng, Aohan and Hou, Lei and others},
  booktitle={Proceedings of the 62nd annual meeting of the association for computational linguistics (volume 1: Long papers)},
  pages={3119--3137},
  year={2024}
}

@article{gu2021efficiently,
  title={Efficiently modeling long sequences with structured state spaces},
  author={Gu, Albert and Goel, Karan and R{\'e}, Christopher},
  journal={arXiv preprint arXiv:2111.00396},
  year={2021}
}

@article{smith2022simplified,
  title={Simplified state space layers for sequence modeling},
  author={Smith, Jimmy TH and Warrington, Andrew and Linderman, Scott W},
  journal={arXiv preprint arXiv:2208.04933},
  year={2022}
}

@inproceedings{orvieto2023resurrecting,
  title={Resurrecting recurrent neural networks for long sequences},
  author={Orvieto, Antonio and Smith, Samuel L and Gu, Albert and Fernando, Anushan and Gulcehre, Caglar and Pascanu, Razvan and De, Soham},
  booktitle={International conference on machine learning},
  pages={26670--26698},
  year={2023},
  organization={PMLR}
}

@article{gardner1988space,
  title={The space of interactions in neural network models},
  author={Gardner, Elizabeth},
  journal={Journal of physics A: Mathematical and general},
  volume={21},
  number={1},
  pages={257--270},
  year={1988}
}

@article{prados1989neural,
  title={Neural network capacity using delta rule},
  author={Prados, DL and Kak, SC},
  journal={Electronics Letters},
  volume={25},
  number={3},
  pages={197--199},
  year={1989},
  publisher={IET}
}

@article{irie2021going,
  title={Going beyond linear transformers with recurrent fast weight programmers},
  author={Irie, Kazuki and Schlag, Imanol and Csord{\'a}s, R{\'o}bert and Schmidhuber, J{\"u}rgen},
  journal={Advances in neural information processing systems},
  volume={34},
  pages={7703--7717},
  year={2021}
}

@inproceedings{schlag2021linear,
  title={Linear transformers are secretly fast weight programmers},
  author={Schlag, Imanol and Irie, Kazuki and Schmidhuber, J{\"u}rgen},
  booktitle={International conference on machine learning},
  pages={9355--9366},
  year={2021},
  organization={PMLR}
}

@article{joffrain2006accumulating,
  title={Accumulating Householder transformations, revisited},
  author={Joffrain, Thierry and Low, Tze Meng and Quintana-Ort{\'\i}, Enrique S and Geijn, Robert van de and Zee, Field G Van},
  journal={ACM Transactions on Mathematical Software (TOMS)},
  volume={32},
  number={2},
  pages={169--179},
  year={2006},
  publisher={ACM New York, NY, USA}
}

@article{karami2025trellis,
  title={Trellis: Learning to compress key-value memory in attention models},
  author={Karami, Mahdi and Behrouz, Ali and Kacham, Praneeth and Mirrokni, Vahab},
  journal={arXiv preprint arXiv:2512.23852},
  year={2025}
}

@misc{eval-harness,
  author       = {Gao, Leo and Tow, Jonathan and Abbasi, Baber and Biderman, Stella and Black, Sid and DiPofi, Anthony and Foster, Charles and Golding, Laurence and Hsu, Jeffrey and Le Noac'h, Alain and Li, Haonan and McDonell, Kyle and Muennighoff, Niklas and Ociepa, Chris and Phang, Jason and Reynolds, Laria and Schoelkopf, Hailey and Skowron, Aviya and Sutawika, Lintang and Tang, Eric and Thite, Anish and Wang, Ben and Wang, Kevin and Zou, Andy},
  title        = {The Language Model Evaluation Harness},
  month        = 07,
  year         = 2024,
  publisher    = {Zenodo},
  version      = {v0.4.3},
  doi          = {10.5281/zenodo.12608602},
  url          = {https://zenodo.org/records/12608602}
}
\bibliographystyle{iclr2027_conference}

\appendix
\section{Appendix}

\subsection{Gated Oja Rule}
\label{app:oja}

Throughout this appendix, we use the state orientation
$\bm{S}_t\in\mathbb{R}^{D_v\times D_k}$, with
$\bm{v}_t,\bm{\alpha}_t\in\mathbb{R}^{D_v}$,
$\bm{k}_t,\bm{q}_t\in\mathbb{R}^{D_k}$, and scalar $\beta_t$.
We assume that the decay entries are strictly positive, so that the
element-wise divisions used below are well defined.

For the gated Oja rule considered here, we first apply a diagonal
value-side decay to the recurrent state, followed by the Oja correction:
\begin{equation}
\label{eq:oja-recurrence}
\begin{aligned}
    \bm{S}_t
    &= {\color{blue}\operatorname{Diag}\left( \bm{\alpha}_t \right)}\bm{S}_{t-1}
    + {\color{blue}\beta_t}\bm{v}_t
    \left(
    \bm{k}_t^\intercal
    -
    \bm{v}_t^\intercal
    {\color{blue}\operatorname{Diag}\left( \bm{\alpha}_t \right)}
    \bm{S}_{t-1}
    \right)
    \\
    &=
    \left(
    \bm{I}
    -
    {\color{blue}\beta_t}\bm{v}_t \bm{v}_t^\intercal
    \right)
    {\color{blue}\operatorname{Diag}\left( \bm{\alpha}_t \right)}
    \bm{S}_{t-1}
    +
    {\color{blue}\beta_t}\bm{v}_t \bm{k}_t^\intercal .
\end{aligned}
\end{equation}

By partially expanding the recurrence, we have
\begin{equation}
\label{eq:oja-expansion}
\begin{aligned}
 \bm{S}_{[t]}^r
 &=
 \underbrace{
 \left(
 \prod_{i=1}^r
 \left(
 \bm{I}
 -
 {\color{blue}\beta_{[t]}^i}
 \bm{v}_{[t]}^i
 \bm{v}_{[t]}^{i\intercal}
 \right)
 {\color{blue}\operatorname{Diag}\left( \bm{\alpha}_{[t]}^i \right)}
 \right)
 }_{:= \bm{D}_{[t]}^r~(\text{``pseudo'' memory decay})}
 \bm{S}_{[t]}^0
 \\
 &\quad+
 \underbrace{
 \sum_{i=1}^{r}
 \left(
 \left(
 \prod_{j=i+1}^{r}
 \left(
 \bm{I}
 -
 {\color{blue}\beta_{[t]}^j}
 \bm{v}_{[t]}^j
 \bm{v}_{[t]}^{j\intercal}
 \right)
 {\color{blue}\operatorname{Diag}\left( \bm{\alpha}_{[t]}^j \right)}
 \right)
 {\color{blue}\beta^i_{[t]}}
 \bm{v}^i_{[t]}
 \bm{k}_{[t]}^{i\intercal}
 \right)
 }_{:= \bm{H}_{[t]}^r~(\text{``pseudo'' incremental memory})}.
\end{aligned}
\end{equation}
where the matrix products are time ordered, with larger time indices appearing
on the left.

For compact notation, we define the element-wise cumulative decay
\[
\bm{\alpha}_{[t]}^{a:b}
:=
\bigodot_{j=a}^{b}\bm{\alpha}_{[t]}^j,
\]
with $\bm{\alpha}_{[t]}^{b+1:b}=\bm{1}$ for an empty product.
We further define the prefix- and suffix-decay matrices
$\bm{A}_{[t]}^{1\rightarrow C},
\bm{A}_{[t]}^{i+1\rightarrow C}\in\mathbb{R}^{C\times D_v}$
row-wise as
\[
\left[\bm{A}_{[t]}^{1\rightarrow C}\right]_{r,:}
=
\bm{\alpha}_{[t]}^{1:r},
\qquad
\left[\bm{A}_{[t]}^{i+1\rightarrow C}\right]_{i,:}
=
\bm{\alpha}_{[t]}^{i+1:C},
\]
for $i,r=1,\ldots,C$.
We also collect the scalar gates within a chunk as
\[
\bm{\beta}_{[t]}
:=
\left[
\beta_{[t]}^1,\ldots,\beta_{[t]}^C
\right]^\intercal .
\]

Then, we employ a WY-style representation \cite{bischof1987wy}:
\begin{equation}
\label{eq:oja-wy-factorization}
\begin{aligned}
\bm{D}_{[t]}^r
&=
{\color{blue}\operatorname{Diag}\left( \bm{\alpha}_{[t]}^{1:r} \right)}
-
\sum_{i=1}^{r}
{\color{blue}\operatorname{Diag}\left( \bm{\alpha}_{[t]}^{i+1:r} \right)}
\bm{v}_{[t]}^i
\bm{w}_{[t]}^{i\intercal},
\\
\bm{H}_{[t]}^r
&=
\sum_{i=1}^r
{\color{blue}\operatorname{Diag}\left( \bm{\alpha}_{[t]}^{i+1:r} \right)}
\bm{v}_{[t]}^i
\bm{u}_{[t]}^{i\intercal}.
\end{aligned}
\end{equation}
The corresponding auxiliary vectors are
\begin{equation}
\label{eq:oja-wy-aux}
\begin{aligned}
\bm{w}_{[t]}^r
&=
{\color{blue}{\beta_{[t]}^{r}}}
\left(
{\color{blue}\operatorname{Diag}\left( \bm{\alpha}_{[t]}^{1:r} \right)}
\bm{v}_{[t]}^r
-
\sum_{i=1}^{r-1}
\bm{w}_{[t]}^i
\left(
\bm{v}_{[t]}^{i\intercal}
{\color{blue}\operatorname{Diag}\left( \bm{\alpha}_{[t]}^{i+1:r} \right)}
\bm{v}_{[t]}^r
\right)
\right),
\\
\bm{u}_{[t]}^r
&=
{\color{blue}\beta_{[t]}^{r}}
\left(
\bm{k}_{[t]}^r
-
\sum_{i=1}^{r-1}
\bm{u}_{[t]}^i
\left(
\bm{v}_{[t]}^{i\intercal}
{\color{blue}\operatorname{Diag}\left( \bm{\alpha}_{[t]}^{i+1:r} \right)}
\bm{v}_{[t]}^r
\right)
\right).
\end{aligned}
\end{equation}
where $\bm{\alpha}_{[t]}^{r+1:r}=\bm{1}$ denotes the empty product.
% \paragraph{UT transform.}

To maximize hardware efficiency, we apply the UT transform
\citep{joffrain2006accumulating} to reduce non-matmul FLOPs, which is crucial
to enable better hardware utilization during training.
\begin{equation}
\label{eq:oja-ut}
\begin{aligned}
\bm{W}_{[t]}
&=
\bm{M}_{[t]}
{\color{blue}\operatorname{Diag}\left(\bm{\beta}_{[t]}\right)}
\left(
{\color{blue}\bm{A}_{[t]}^{1\rightarrow C}}
\odot
\bm{V}_{[t]}
\right),
\quad\quad
\bm{U}_{[t]}
=
\bm{M}_{[t]}
{\color{blue}\operatorname{Diag}\left(\bm{\beta}_{[t]}\right)}
\bm{K}_{[t]},
\\
\bm{M}_{[t]}
&=
\left(
\bm{I}
+
\operatorname{lower}
\left(
{\color{blue}\operatorname{Diag}\left(\bm{\beta}_{[t]}\right)}
\left(
\bm{V}_{[t]}
\odot
{\color{blue}\bm{A}_{[t]}^{1\rightarrow C}}
\right)
\left(
\bm{V}_{[t]}
\oslash
{\color{blue}\bm{A}_{[t]}^{1\rightarrow C}}
\right)^\intercal
\right)
\right)^{-1}.
\end{aligned}
\end{equation}
where
$\bm{M}_{[t]}\in\mathbb{R}^{C\times C}$,
$\bm{W}_{[t]}\in\mathbb{R}^{C\times D_v}$,
$\bm{U}_{[t]}\in\mathbb{R}^{C\times D_k}$, and
$\operatorname{lower}(\cdot)$ extracts the strictly lower-triangular part.
% Note that in Gated Deltanet, ${\color{blue}{\mathcal{A}}}$ is a skew-symmetric matrix, so only the negative part can be retained to stabilize training. However, this does not apply to ${\color{blue}\tilde{\mathcal{A}}_{ij}=\bm{\alpha}^{1:i-1}/\bm{\alpha}^{1:j}}$.

The corresponding lower-triangular systems can be efficiently solved through
an iterative row-wise approach using forward substitution
, with the computation maintained in
$\operatorname{float32}$.
% ${\color{blue}\bar{\mathcal{A}}_{[t]}^{(i-1)/j}=\bm{\alpha}_{[t]}^{1:i}/(\bm{\alpha}_{[t]}^{1:j} *\bm{\alpha}_{[t]}^{i})}$.

Then we have the following vector form:
\begin{equation}
\label{eq:oja-vector}
\begin{aligned}
\bm{S}_{[t]}^r
&=
\bm{D}_{[t]}^r \bm{S}_{[t]}^0
+
\bm{H}_{[t]}^r
\\
&=
{\color{blue}\operatorname{Diag}\left( \bm{\alpha}_{[t]}^{1:r} \right)}
\bm{S}_{[t]}^0
+
\sum_{i=1}^r
{\color{blue}\operatorname{Diag}\left( \bm{\alpha}_{[t]}^{i+1:r} \right)}
\bm{v}_{[t]}^i
\left(
\bm{u}_{[t]}^{i\intercal}
-
\left(
\bm{w}_{[t]}^{i\intercal}
\bm{S}_{[t]}^0
\right)
\right),
\\
\bm{o}_{[t]}^r
&=
\bm{S}_{[t]}^r
\bm{q}_{[t]}^r
\\
&=
{\color{blue}\operatorname{Diag}\left( \bm{\alpha}_{[t]}^{1:r} \right)}
\bm{S}_{[t]}^0
\bm{q}_{[t]}^r
+
\sum_{i=1}^r
{\color{blue}\operatorname{Diag}\left( \bm{\alpha}_{[t]}^{i+1:r} \right)}
\bm{v}_{[t]}^i
\left(
\bm{u}_{[t]}^{i\intercal}
-
\left(
\bm{w}_{[t]}^{i\intercal}
\bm{S}_{[t]}^0
\right)
\right)
\bm{q}_{[t]}^r .
\end{aligned}
\end{equation}

Within chunk $[t]$, we identify
$\bm{S}_{[t]}^0=\bm{S}_{[t]}$ and
$\bm{S}_{[t]}^C=\bm{S}_{[t+1]}$.
Equivalently, in matrix form:
\begin{equation}
\label{eq:oja-matrix}
\begin{aligned}
\bm{S}_{[t+1]}
&=
{\color{blue}\operatorname{Diag}\left( \bm{\alpha}_{[t]}^{1:C} \right)}
\bm{S}_{[t]}
+
\left(
{\color{blue}\bm{A}_{[t]}^{i+1\rightarrow C}}
\odot
\bm{V}_{[t]}
\right)^\intercal
\left(
\bm{U}_{[t]}
-
\bm{W}_{[t]}
\bm{S}_{[t]}
\right),
\\
\bm{O}_{[t]}
&=
{\color{blue}\bm{A}_{[t]}^{1\rightarrow C}}
\odot
\left(
\bm{Q}_{[t]}
\bm{S}_{[t]}^\intercal
+
\operatorname{Tril}
\left(
\bm{Q}_{[t]}
\underbrace{
\left(
\bm{U}_{[t]}
-
\bm{W}_{[t]}
\bm{S}_{[t]}
\right)^\intercal
}_{\text{``pseudo''-key term}}
\right)
\left(
\bm{V}_{[t]}
\oslash
{\color{blue}\bm{A}_{[t]}^{1\rightarrow C}}
\right)
\right).
\end{aligned}
\end{equation}
Here, $\operatorname{Tril}(\cdot)$ retains the lower-triangular part including
the diagonal.

\subsection{Gated Oja Rule-2}
\label{app:oja2}

Inspired by the erase--write decoupling in GDN2~\citep{hatamizadeh2026gated},
Gated Oja Rule-2 replaces the shared scalar gate with a value-side erase gate
$\bm{b}_t\in\mathbb{R}^{D_v}$ and a channel-wise write gate
$\bm{c}_t\in\mathbb{R}^{D_k}$, respectively.
Under our $D_v\times D_k$ state orientation, the Oja correction is applied
through the value-side direction $\bm{v}_t$; hence the erase gate modulates
$\bm{v}_t$, whereas the write gate modulates $\bm{k}_t$:
\begin{equation}
\label{eq:oja2-recurrence}
\begin{aligned}
    \bm{S}_t
    &=
    {\color{blue}\operatorname{Diag}\left( \bm{\alpha}_t \right)}
    \bm{S}_{t-1}
    +
    \bm{v}_t
    \left(
    ({\color{blue}\bm{c}_t}\odot\bm{k}_t)^\intercal
    -
    ({\color{blue}\bm{b}_t}\odot\bm{v}_t)^\intercal
    {\color{blue}\operatorname{Diag}\left( \bm{\alpha}_t \right)}
    \bm{S}_{t-1}
    \right)
    \\
    &=
    \left(
    \bm{I}
    -
    \bm{v}_t
    ({\color{blue}\bm{b}_t}\odot\bm{v}_t)^\intercal
    \right)
    {\color{blue}\operatorname{Diag}\left( \bm{\alpha}_t \right)}
    \bm{S}_{t-1}
    +
    \bm{v}_t
    ({\color{blue}\bm{c}_t}\odot\bm{k}_t)^\intercal .
\end{aligned}
\end{equation}

By partially expanding the recurrence, we have
\begin{equation}
\label{eq:oja2-expansion}
\begin{aligned}
 \bm{S}_{[t]}^r
 &=
 \underbrace{
 \left(
 \prod_{i=1}^r
 \left(
 \bm{I}
 -
 \bm{v}_{[t]}^i
 ({\color{blue}\bm{b}_{[t]}^i}\odot\bm{v}_{[t]}^{i})^{\intercal}
 \right)
 {\color{blue}\operatorname{Diag}\left( \bm{\alpha}_{[t]}^i \right)}
 \right)
 }_{:= \bm{D}_{[t]}^r~(\text{``pseudo'' memory decay})}
 \bm{S}_{[t]}^0
 \\
 &\quad+
 \underbrace{
 \sum_{i=1}^{r}
 \left(
 \left(
 \prod_{j=i+1}^{r}
 \left(
 \bm{I}
 -
 \bm{v}_{[t]}^j
 ({\color{blue}\bm{b}_{[t]}^j}\odot\bm{v}_{[t]}^{j})^{\intercal}
 \right)
 {\color{blue}\operatorname{Diag}\left( \bm{\alpha}_{[t]}^j \right)}
 \right)
 \bm{v}^i_{[t]}
 ({\color{blue}\bm{c}^i_{[t]}}\odot\bm{k}_{[t]}^{i})^{\intercal}
 \right)
 }_{:= \bm{H}_{[t]}^r~(\text{``pseudo'' incremental memory})}.
\end{aligned}
\end{equation}
where the matrix products are time ordered, with larger time indices
appearing on the left.

We retain the same WY factorization structure of $\bm{D}_{[t]}^r$ and
$\bm{H}_{[t]}^r$ as in Eq.~\eqref{eq:oja-wy-factorization}. Under the
decoupled erase and write gates, however, the auxiliary recurrences in
Eq.~\eqref{eq:oja-wy-aux} are replaced by
\begin{equation}
\label{eq:oja2-wy-aux}
\begin{aligned}
\bm{w}_{[t]}^r
&=
{\color{blue}\operatorname{Diag}\left(
\bm{\alpha}_{[t]}^{1:r}
\right)}
({\color{blue}\bm{b}_{[t]}^{r}}\odot\bm{v}_{[t]}^r)
-
\sum_{i=1}^{r-1}
\bm{w}_{[t]}^i
\left(
\bm{v}_{[t]}^{i\intercal}
{\color{blue}\operatorname{Diag}\left(
\bm{\alpha}_{[t]}^{i+1:r}
\right)}
({\color{blue}\bm{b}_{[t]}^{r}}\odot\bm{v}_{[t]}^r)
\right),
\\
\bm{u}_{[t]}^r
&=
{\color{blue}\bm{c}_{[t]}^{r}}\odot\bm{k}_{[t]}^r
-
\sum_{i=1}^{r-1}
\bm{u}_{[t]}^i
\left(
\bm{v}_{[t]}^{i\intercal}
{\color{blue}\operatorname{Diag}\left(
\bm{\alpha}_{[t]}^{i+1:r}
\right)}
({\color{blue}\bm{b}_{[t]}^{r}}\odot\bm{v}_{[t]}^r)
\right).
\end{aligned}
\end{equation}
where $\bm{\alpha}_{[t]}^{r+1:r}=\bm{1}$ denotes the empty product.

We collect the erase and write gates within a chunk as
\[
\bm{B}_{[t]}
=
\begin{bmatrix}
(\bm{b}_{[t]}^1)^\intercal\\
\vdots\\
(\bm{b}_{[t]}^C)^\intercal
\end{bmatrix}
\in\mathbb{R}^{C\times D_v},
\qquad
\bm{C}_{[t]}
=
\begin{bmatrix}
(\bm{c}_{[t]}^1)^\intercal\\
\vdots\\
(\bm{c}_{[t]}^C)^\intercal
\end{bmatrix}
\in\mathbb{R}^{C\times D_k}.
\]

To maximize hardware efficiency, we apply the UT transform
\citep{joffrain2006accumulating} to the gate-aware WY recurrences above,
reducing non-matmul FLOPs and enabling efficient chunkwise computation:
\begin{equation}
\label{eq:oja2-ut}
\begin{aligned}
\bm{W}_{[t]}
&=
\bm{M}_{[t]}
\left(
{\color{blue}\bm{A}_{[t]}^{1\rightarrow C}}
\odot
({\color{blue}\bm{B}_{[t]}}\odot\bm{V}_{[t]})
\right),
\quad\quad
\bm{U}_{[t]}
=
\bm{M}_{[t]}
({\color{blue}\bm{C}_{[t]}}\odot\bm{K}_{[t]}),
\\
\bm{M}_{[t]}
&=
\left(
\bm{I}
+
\operatorname{lower}
\left(
\left(
({\color{blue}\bm{B}_{[t]}}\odot\bm{V}_{[t]})
\odot
{\color{blue}\bm{A}_{[t]}^{1\rightarrow C}}
\right)
\left(
\bm{V}_{[t]}
\oslash
{\color{blue}\bm{A}_{[t]}^{1\rightarrow C}}
\right)^\intercal
\right)
\right)^{-1}.
\end{aligned}
\end{equation}
where $\operatorname{lower}(\cdot)$ extracts the strictly lower-triangular
part.

Unlike the Gated Oja Rule, the erase and write operations are controlled by
separate gates, so there is no shared gate factor
$\operatorname{Diag}(\bm{\beta}_{[t]})$ that can be factored out from both
branches. Instead, the erase gate $\bm{B}_{[t]}$ is absorbed into the
correction-side factors and therefore affects both $\bm{M}_{[t]}$ and
$\bm{W}_{[t]}$, whereas the write gate $\bm{C}_{[t]}$ acts only on the
write-side factor $\bm{U}_{[t]}$. With the modified auxiliary recurrences in
Eq.~\eqref{eq:oja2-wy-aux} and the UT factors in Eq.~\eqref{eq:oja2-ut}, the
resulting vector and matrix forms retain the same structure as
Eqs.~\eqref{eq:oja-vector} and~\eqref{eq:oja-matrix}, respectively.

\subsection{Experimental Details}
\label{app:exp}
\subsubsection{Training Details}
\label{app:training_details}

All models trained in this work contain approximately 1.3B parameters
and are trained on 100B tokens from FineWeb-Edu~\citep{penedo2024fineweb}.
We use the TinyLlama v1.1 tokenizer, which follows the Llama-style
SentencePiece BPE vocabulary with 32,000 tokens.
No additional padding tokens are introduced, since the vocabulary size
is already a multiple of 64.

We train all models using AdamW with
$\beta_1=0.9$ and $\beta_2=0.95$.
The peak learning rate is $4\times10^{-4}$, with weight decay of 0.1
and gradient clipping at 1.0.
The learning rate is linearly warmed up over the first 1B training
tokens and then decayed with a cosine schedule to 10\% of the peak
learning rate.
The global batch size is approximately 0.5M tokens,
corresponding to 128 sequences of length 4,096.
Training uses BF16 mixed precision.
All models are trained on four compute nodes with eight NVIDIA H20 GPUs
per node, for a total of 32 GPUs, using CUDA 12.8.

The training sequence length is 4K tokens.
For hybrid models, the recurrent mixer is interleaved with
sliding-window attention using a 2K attention window.
Unless otherwise stated, all models trained in this work follow the
same training recipe.

\subsubsection{Benchmarks Details}
\label{app:evaluation}

\paragraph{Language modeling and commonsense reasoning.}
We evaluate language modeling with perplexity on WikiText
and LAMBADA.
For zero-shot evaluation, we report LAMBADA accuracy together with
PIQA, HellaSwag, WinoGrande, ARC-Easy, ARC-Challenge,
OpenBookQA, Social IQa, and BoolQ.
The commonsense benchmarks are evaluated using
\texttt{lm-evaluation-harness}~\citep{eval-harness}.

\paragraph{Synthetic long-context retrieval.}
We evaluate long-context retrieval using RULER~\citep{hsieh2024ruler}.
The needle-in-a-haystack evaluation includes the S-NIAH and
MK-NIAH task groups at multiple context lengths.
We additionally evaluate the remaining RULER tasks, including
multi-query (MQ), multi-value (MV), common-word extraction (CWE),
frequent-word extraction (FWE), HotpotQA (HPQA), SQuAD, and
variable tracking (VT).
We follow the official RULER evaluation protocol.

\paragraph{Real-world retrieval.}
Following~\citet{arora2024just}, we evaluate real-world retrieval
on SWDE, SQuAD, FDA, TriviaQA, Natural Questions (NQ), and DROP.
All inputs are truncated to 2K tokens.

\paragraph{Long-context understanding.}
We evaluate general long-context understanding on 11 tasks from
LongBench~\citep{bai2024longbench}.
These include MultiFieldQA and Qasper for single-document QA;
HotpotQA and 2WikiMultiHopQA for multi-document QA;
GovReport and MultiNews for summarization;
SAMSum, TREC, and TriviaQA for few-shot evaluation;
and LCC and RepoBench-P for code.
We report the average over all 11 tasks.

\subsubsection{Training Throughput Evaluation}
\label{app:training_efficiency}

We benchmark end-to-end training throughput on a single NVIDIA H20 GPU
under a fixed budget of 16,384 tokens per optimization step.
We evaluate sequence lengths of 2K, 4K, 8K, and 16K with corresponding
batch sizes of 8, 4, 2, and 1, respectively, without gradient accumulation.
All compared models contain approximately 1.30B trainable parameters
and use the same BF16 mixed-precision configuration.

Each measured iteration consists of a complete optimization step,
including gradient clearing, forward propagation, fused cross-entropy,
backward propagation, gradient clipping, and a fused AdamW update.
To isolate model computation from the input pipeline, token IDs are
sampled uniformly and generated directly on the GPU, excluding
tokenization, storage, and data-loading overhead.

For each configuration, we discard 20 warm-up iterations and measure
the subsequent 50 iterations.
We repeat each benchmark on three different physical H20 GPUs,
compute the mean throughput within each run, and report the median
of the three run-level means.
Throughput is measured in thousands of processed tokens per second
(Kt/s).

\begin{table*}[t!]
\centering
% \scriptsize
\setlength{\tabcolsep}{1.45pt}
\renewcommand{\arraystretch}{1.05}

\begin{adjustbox}{width=\textwidth,center}

\begin{tabular}{
@{}l
*{24}{c}
>{\columncolor{red!7}}c
@{}
}

\toprule

\multirow{2}{*}{\textbf{Model}}
& \multicolumn{4}{c}{\textbf{S-NIAH-1}}
& \multicolumn{4}{c}{\textbf{S-NIAH-2}}
& \multicolumn{4}{c}{\textbf{S-NIAH-3}}
& \multicolumn{4}{c}{\textbf{MK-NIAH-1}}
& \multicolumn{4}{c}{\textbf{MK-NIAH-2}}
& \multicolumn{4}{c}{\textbf{MK-NIAH-3}}
& \textbf{Avg.}
\\

\cmidrule(lr){2-5}
\cmidrule(lr){6-9}
\cmidrule(lr){10-13}
\cmidrule(lr){14-17}
\cmidrule(lr){18-21}
\cmidrule(lr){22-25}

& \textbf{1K} & \textbf{2K} & \textbf{4K} & \textbf{8K}
& \textbf{1K} & \textbf{2K} & \textbf{4K} & \textbf{8K}
& \textbf{1K} & \textbf{2K} & \textbf{4K} & \textbf{8K}
& \textbf{1K} & \textbf{2K} & \textbf{4K} & \textbf{8K}
& \textbf{1K} & \textbf{2K} & \textbf{4K} & \textbf{8K}
& \textbf{1K} & \textbf{2K} & \textbf{4K} & \textbf{8K}
& $\uparrow$
\\

\midrule

\multicolumn{26}{l}{\textit{Recurrent models}} \\

\rulermodel{Mamba-2}
& 68.6 & 62.4 & 41.6 & 51.0
& 52.8 & 73.6 & 23.0 & 9.8
& 0.0 & 0.0 & 10.6 & 4.6
& 24.8 & 18.6 & 14.0 & 7.2
& 0.2 & 0.6 & 0.0 & \rulersecond{0.0}
& \rulerbest{0.0} & \rulerbest{0.0}
& \rulerbest{0.0} & \rulerbest{0.0}
& 19.3
\\

\rulermodel{Gated DeltaNet}
& \rulerbest{100.0} & \rulerbest{100.0} & 94.6 & 46.0
& \rulerbest{100.0} & 97.2 & 60.2 & 22.6
& \rulersecond{84.6} & \rulerbest{69.0}
& \rulerbest{37.2} & 6.6
& 41.6 & 37.8 & \rulersecond{28.2} & 21.0
& 2.4 & 0.6 & 0.0 & \rulersecond{0.0}
& \rulerbest{0.0} & \rulerbest{0.0}
& \rulerbest{0.0} & \rulerbest{0.0}
& 39.6
\\

\rulermodel{KDA}
& \rulerbest{100.0} & \rulerbest{100.0}
& \rulerbest{100.0} & \rulersecond{90.0}
& \rulerbest{100.0} & 95.6 & 50.2 & 20.8
& 84.0 & 57.2 & \rulersecond{31.2} & \rulersecond{9.6}
& 49.6 & 34.8 & 16.6 & 13.6
& 2.4 & 0.4 & 0.0 & \rulersecond{0.0}
& \rulerbest{0.0} & \rulerbest{0.0}
& \rulerbest{0.0} & \rulerbest{0.0}
& \rulersecond{39.8}
\\

\rulermodel{Mamba-3 (SISO)}
& \rulerbest{100.0} & 98.0 & 82.0 & 37.8
& \rulerbest{100.0} & 96.6 & 37.8 & 14.2
& 80.4 & 43.8 & 14.0 & 3.8
& \rulersecond{60.6} & 37.0 & 25.6 & 12.8
& \rulersecond{3.6} & \rulerbest{1.6}
& \rulerbest{0.4} & \rulersecond{0.0}
& \rulerbest{0.0} & \rulerbest{0.0}
& \rulerbest{0.0} & \rulerbest{0.0}
& 35.4
\\

\rulermodel{Mamba-3 (MIMO)}
& \rulerbest{100.0} & \rulersecond{99.8} & 98.4 & 69.4
& \rulerbest{100.0} & \rulersecond{98.6} & 55.0 & 15.6
& 61.8 & 24.8 & 11.8 & 2.4
& \rulerbest{65.8} & \rulerbest{55.8} & 25.8 & 11.8
& \rulerbest{5.0} & \rulersecond{1.0}
& \rulersecond{0.2} & \rulersecond{0.0}
& \rulerbest{0.0} & \rulerbest{0.0}
& \rulerbest{0.0} & \rulerbest{0.0}
& 37.6
\\

\rulermodel{Gated DeltaNet-2}
& \rulerbest{100.0} & \rulerbest{100.0}
& \rulerbest{100.0} & \rulerbest{92.8}
& \rulersecond{99.8} & 87.2
& \rulersecond{60.6} & \rulersecond{23.8}
& 78.4 & 41.2 & 26.4 & \rulerbest{12.8}
& 40.8 & 31.8 & \rulerbest{29.6} & \rulerbest{25.8}
& 1.2 & 0.2 & \rulersecond{0.2} & \rulerbest{0.2}
& \rulerbest{0.0} & \rulerbest{0.0}
& \rulerbest{0.0} & \rulerbest{0.0}
& 39.7
\\

\rulermodel{OJA}
& \rulerbest{100.0} & 90.2 & 61.6 & 23.2
& \rulersecond{99.8} & 96.4 & 31.6 & 17.8
& 67.0 & 24.2 & 10.6 & 4.8
& 25.6 & 16.4 & 16.8 & 11.8
& 0.0 & 0.0 & 0.0 & \rulersecond{0.0}
& \rulerbest{0.0} & \rulerbest{0.0}
& \rulerbest{0.0} & \rulerbest{0.0}
& 29.1
\\

\rulermodel{OJA2}
& \rulersecond{99.8} & 92.4 & 54.0 & 12.8
& \rulerbest{100.0} & 95.2 & 50.6 & 20.0
& 70.8 & 34.4 & 20.0 & 3.4
& 33.2 & 20.6 & 18.6 & 11.4
& 0.4 & 0.0 & 0.0 & \rulersecond{0.0}
& \rulerbest{0.0} & \rulerbest{0.0}
& \rulerbest{0.0} & \rulerbest{0.0}
& 30.7
\\

\rulermodel{Gated Slot Attention}
& 98.4 & 78.4 & 39.8 & 18.2
& 99.0 & 83.2 & 19.6 & 6.4
& 56.2 & 20.4 & 7.2 & 1.4
& 27.8 & 22.0 & 18.8 & 7.4
& 0.6 & 0.4 & 0.0 & \rulersecond{0.0}
& \rulerbest{0.0} & \rulerbest{0.0}
& \rulerbest{0.0} & \rulerbest{0.0}
& 25.2
\\

\rowcolor{blue!10}
\rulermodel{Gated Slot Attention-2}
& \rulerbest{100.0} & \rulerbest{100.0}
& \rulersecond{99.6} & 86.0
& \rulerbest{100.0} & \rulerbest{99.2}
& \rulerbest{70.4} & \rulerbest{30.2}
& \rulerbest{92.0} & \rulersecond{59.6}
& 27.0 & \rulersecond{9.6}
& 45.2 & \rulersecond{39.8} & 26.0 & \rulersecond{23.2}
& 1.8 & 0.6 & 0.0 & \rulerbest{0.2}
& \rulerbest{0.0} & \rulerbest{0.0}
& \rulerbest{0.0} & \rulerbest{0.0}
& \rulerbest{42.1}
\\

\midrule

\multicolumn{26}{l}{\textit{Attention or hybrid models}} \\

\rowcolor{gray!30}
\rulermodel{Transformer}
& \rulerbest{100.0} & \rulerbest{100.0}
& \rulerbest{99.6} & 0.0
& \rulerbest{100.0} & \rulerbest{100.0}
& \rulerbest{94.6} & 0.0
& 79.2 & 65.4 & \rulerbest{61.2} & 0.0
& 80.2 & 81.0 & \rulerbest{65.8} & 0.0
& 15.8 & 11.2 & 6.4 & 0.0
& 25.4 & \rulerbest{18.8}
& \rulerbest{6.2} & 0.0
& 46.3
\\

\rulermodel{Mamba-2}
& 70.6 & 59.8 & 24.6 & 0.0
& 45.2 & 87.2 & 24.4 & 0.0
& 0.8 & 0.2 & 45.2 & 0.0
& 41.8 & 38.0 & 23.8 & 0.0
& 11.6 & 11.8 & 1.4 & 0.0
& 0.2 & 0.0 & 0.0 & 0.0
& 20.3
\\

\rulermodel{Gated DeltaNet}
& \rulersecond{99.8} & 98.2 & 47.8 & 20.2
& \rulersecond{99.8} & \rulersecond{99.8}
& 51.0 & 0.8
& 96.0 & 87.2 & 55.6 & 0.0
& 83.0 & \rulersecond{85.0} & 47.2 & 1.6
& 40.6 & 19.6 & 3.2 & 1.2
& 21.0 & 8.4 & 0.2 & 0.6
& 44.5
\\

\rulermodel{KDA}
& 99.6 & \rulersecond{99.8} & 53.2 & 24.0
& \rulersecond{99.8} & \rulersecond{99.8}
& 48.8 & 16.8
& 86.6 & 65.2 & 38.8 & 3.8
& 78.6 & 68.6 & 40.8 & 13.0
& \rulersecond{58.2} & 32.2 & 4.4 & 0.2
& 26.6 & 10.4 & 2.4 & 0.2
& 44.7
\\

\rulermodel{Mamba-3 (SISO)}
& \rulerbest{100.0} & \rulerbest{100.0}
& \rulersecond{53.8} & 2.2
& \rulerbest{100.0} & \rulerbest{100.0}
& 51.0 & 0.0
& 86.2 & 78.8 & 52.8 & 0.0
& 63.8 & 54.6 & 41.6 & 0.0
& 11.8 & 6.8 & 2.0 & 0.0
& 10.8 & 4.8 & 1.2 & 0.0
& 38.4
\\

\rulermodel{Mamba-3 (MIMO)}
& \rulerbest{100.0} & \rulerbest{100.0}
& \rulersecond{53.8} & \rulerbest{25.2}
& \rulerbest{100.0} & \rulerbest{100.0}
& 51.0 & 0.4
& \rulerbest{98.8} & \rulerbest{98.8}
& 56.4 & 0.6
& 62.8 & 68.6 & 45.0 & 3.8
& 25.6 & 12.0 & 4.8 & 0.2
& 18.2 & 6.8 & 0.8 & 0.2
& 43.1
\\

\rulermodel{Gated DeltaNet-2}
& \rulerbest{100.0} & 92.6 & 48.2 & 16.2
& \rulerbest{100.0} & \rulerbest{100.0}
& \rulersecond{51.6} & 16.2
& 96.2 & 80.6 & 47.6 & 9.4
& \rulersecond{88.6} & 83.0 & 45.8 & 13.4
& 55.4 & \rulersecond{50.2}
& \rulerbest{20.0} & 0.2
& 21.8 & 5.8 & 1.8 & 0.0
& \rulersecond{47.7}
\\

\rulermodel{OJA}
& \rulerbest{100.0} & \rulerbest{100.0}
& 51.8 & 19.2
& \rulerbest{100.0} & 99.0
& 50.8 & \rulerbest{19.4}
& 87.6 & 60.4 & 38.6 & \rulersecond{9.8}
& 72.8 & 57.2 & 49.2 & \rulersecond{18.2}
& 22.2 & 8.6 & 1.6 & 0.2
& 32.6 & 9.6 & \rulersecond{5.8} & \rulersecond{0.8}
& 42.3
\\

\rulermodel{OJA2}
& \rulerbest{100.0} & \rulerbest{100.0}
& 53.6 & \rulerbest{25.2}
& \rulerbest{100.0} & \rulerbest{100.0}
& 51.0 & 13.6
& \rulersecond{97.4} & 81.8 & 54.4 & \rulerbest{12.4}
& 83.8 & 68.4 & 44.8 & 14.8
& 26.4 & 11.2 & 1.0 & \rulersecond{1.8}
& \rulerbest{46.8} & 13.2 & 5.6 & \rulerbest{1.2}
& 46.2
\\

\rulermodel{Gated Slot Attention}
& \rulerbest{100.0} & \rulerbest{100.0}
& \rulersecond{53.8} & 0.0
& \rulerbest{100.0} & \rulerbest{100.0}
& 51.0 & 0.2
& 97.2 & \rulersecond{94.2}
& \rulersecond{56.8} & 0.0
& 69.2 & 67.8 & 54.0 & 0.6
& 41.0 & 34.4 & 11.2 & 0.0
& 30.4 & 10.6 & 0.8 & 0.0
& 44.7
\\

\rowcolor{blue!10}
\rulermodel{Gated Slot Attention-2}
& \rulerbest{100.0} & 99.6 & 52.6
& \rulersecond{24.6}
& \rulerbest{100.0} & \rulerbest{100.0}
& 50.8 & \rulersecond{18.0}
& 86.0 & 70.6 & 36.4 & 7.0
& \rulerbest{92.4} & \rulerbest{94.8}
& \rulersecond{57.4} & \rulerbest{20.0}
& \rulerbest{77.4} & \rulerbest{53.0}
& \rulersecond{14.0} & \rulerbest{4.0}
& \rulersecond{32.8} & \rulersecond{16.6}
& \rulersecond{5.8} & \rulersecond{0.8}
& \rulerbest{50.6}
\\

\bottomrule

\end{tabular}
\end{adjustbox}

\caption{
Accuracy on RULER single-needle (S-NIAH) and multi-key needle-in-a-haystack
(MK-NIAH) tasks across context lengths.
Avg. is computed over all 24 settings.
All non-Transformer models in the attention/hybrid group use SWA.
Best and second-best results within each model family are bolded and underlined, respectively.
}
\label{tab:ruler_niah}

\end{table*}

\begin{table*}[t!]
\centering
% \scriptsize
\setlength{\tabcolsep}{1.15pt}
\renewcommand{\arraystretch}{1.05}

\begin{adjustbox}{width=\textwidth,center}

\begin{tabular}{
@{}l
*{28}{c}
>{\columncolor{red!7}}c
@{}
}

\toprule

\multirow{2}{*}{\textbf{Model}}
& \multicolumn{4}{c}{\textbf{MQ}}
& \multicolumn{4}{c}{\textbf{MV}}
& \multicolumn{4}{c}{\textbf{CWE}}
& \multicolumn{4}{c}{\textbf{FWE}}
& \multicolumn{4}{c}{\textbf{HPQA}}
& \multicolumn{4}{c}{\textbf{SQuAD}}
& \multicolumn{4}{c}{\textbf{VT}}
& \textbf{Avg.}
\\

\cmidrule(lr){2-5}
\cmidrule(lr){6-9}
\cmidrule(lr){10-13}
\cmidrule(lr){14-17}
\cmidrule(lr){18-21}
\cmidrule(lr){22-25}
\cmidrule(lr){26-29}

& \textbf{1K} & \textbf{2K} & \textbf{4K} & \textbf{8K}
& \textbf{1K} & \textbf{2K} & \textbf{4K} & \textbf{8K}
& \textbf{1K} & \textbf{2K} & \textbf{4K} & \textbf{8K}
& \textbf{1K} & \textbf{2K} & \textbf{4K} & \textbf{8K}
& \textbf{1K} & \textbf{2K} & \textbf{4K} & \textbf{8K}
& \textbf{1K} & \textbf{2K} & \textbf{4K} & \textbf{8K}
& \textbf{1K} & \textbf{2K} & \textbf{4K} & \textbf{8K}
& $\uparrow$
\\

\midrule

\multicolumn{30}{l}{\textit{Recurrent models}} \\

\rulermodel{Mamba-2}
& 21.3 & 21.4 & 14.3 & 3.3
& 28.1 & 27.1 & 15.3 & 5.5
& 17.9 & 8.4 & 13.6 & 8.3
& 22.1 & 15.1 & 9.4 & 3.7
& 19.0 & 13.2 & 10.8 & 11.2
& 17.8 & 15.8 & 11.8 & 8.2
& 24.8 & 15.8 & 18.5 & 0.2
& 14.4
\\

\rulermodel{Gated DeltaNet}
& 79.7 & 67.7 & \rulerbest{33.0} & \rulerbest{13.9}
& \rulerbest{76.5} & 49.4 & \rulersecond{29.4} & \rulerbest{17.1}
& \rulersecond{58.1} & 20.3 & \rulersecond{24.7} & 8.9
& 40.2 & \rulersecond{41.2} & 32.3 & 19.5
& 20.4 & 16.0 & 12.6 & 14.2
& 26.6 & 17.6 & 15.2 & 12.6
& 4.5 & 5.9 & 14.2 & 6.1
& 27.8
\\

\rulermodel{KDA}
& \rulerbest{88.9} & \rulersecond{70.6} & 15.7 & 4.9
& 70.8 & 60.2 & 26.6 & 8.3
& 41.3 & 11.3 & 23.4 & \rulersecond{14.1}
& 37.4 & 38.1 & 34.1 & 30.5
& \rulersecond{23.0} & 14.8 & 13.6 & \rulerbest{15.2}
& \rulersecond{30.2} & \rulersecond{21.6}
& \rulerbest{17.8} & \rulerbest{16.4}
& 7.4 & 3.2 & 6.4 & \rulersecond{11.2}
& 27.0
\\

\rulermodel{Mamba-3 (SISO)}
& 85.3 & 69.7 & 28.1 & 6.3
& 70.8 & \rulersecond{64.7} & 27.9 & 10.4
& 17.6 & 12.3 & 11.2 & 1.5
& 40.0 & 38.2 & \rulerbest{40.4} & \rulerbest{34.5}
& 21.4 & \rulersecond{17.2} & 10.8 & 13.0
& 27.6 & 20.6 & 16.4 & \rulersecond{15.4}
& 3.8 & 4.0 & 8.2 & 2.9
& 25.7
\\

\rulermodel{Mamba-3 (MIMO)}
& 85.0 & 67.3 & 24.9 & 7.0
& 73.2 & \rulerbest{65.8} & 29.2 & 8.9
& 36.7 & 15.4 & 23.9 & 6.8
& 39.8 & 32.5 & 30.3 & 24.7
& 22.4 & 15.6 & \rulersecond{14.2} & \rulersecond{15.0}
& 27.0 & 20.6 & \rulerbest{17.8} & 14.2
& \rulerbest{36.2} & \rulersecond{22.6}
& \rulerbest{20.3} & 1.8
& \rulersecond{28.5}
\\

\rulermodel{Gated DeltaNet-2}
& \rulersecond{86.1} & \rulerbest{75.0}
& \rulersecond{32.6} & 7.1
& 64.2 & 60.5 & 23.8 & 5.4
& 33.9 & 15.4 & 23.5 & 13.6
& 24.1 & 26.6 & 17.6 & 18.3
& \rulerbest{27.0} & \rulerbest{18.6}
& \rulerbest{14.8} & \rulersecond{15.0}
& \rulerbest{30.6} & \rulerbest{22.8}
& \rulersecond{16.8} & 14.6
& 8.8 & 6.2 & 9.5 & \rulerbest{15.4}
& 26.0
\\

\rulermodel{OJA}
& 65.5 & 37.7 & 16.1 & 6.3
& 49.0 & 38.5 & 17.1 & 11.2
& 10.9 & 5.8 & 10.0 & 1.8
& 22.3 & 19.3 & 11.8 & 11.9
& 20.2 & 15.0 & 9.2 & 12.4
& 21.2 & 16.2 & 12.6 & 9.4
& 1.6 & 1.3 & 0.0 & 0.0
& 16.2
\\

\rulermodel{OJA2}
& 59.4 & 32.6 & 13.2 & 4.8
& 44.5 & 26.7 & 10.7 & 5.1
& 13.8 & 5.3 & 11.8 & 4.8
& 12.6 & 16.2 & 11.6 & 6.0
& 19.0 & 14.2 & 11.0 & 13.8
& 17.4 & 14.8 & 13.4 & 12.2
& 9.3 & 1.4 & 1.0 & 0.1
& 14.5
\\

\rulermodel{Gated Slot Attention}
& 48.4 & 33.9 & 15.7 & 2.1
& 39.6 & 29.9 & 14.7 & 3.4
& 37.1 & \rulersecond{21.4} & 10.4 & 10.6
& \rulersecond{40.6} & 29.9 & 25.2 & 15.1
& 17.2 & 12.6 & 11.4 & 13.0
& 16.0 & 11.2 & 7.4 & 7.4
& 25.0 & 3.0 & 0.1 & 0.0
& 17.9
\\

\rowcolor{blue!10}
\rulermodel{Gated Slot Attention-2}
& 77.0 & 57.4 & 30.6 & \rulersecond{10.4}
& \rulersecond{73.3} & 60.9
& \rulerbest{43.6} & \rulersecond{14.7}
& \rulerbest{66.3} & \rulerbest{41.1}
& \rulerbest{40.7} & \rulerbest{19.5}
& \rulerbest{42.5} & \rulerbest{44.0}
& \rulersecond{39.3} & \rulersecond{33.0}
& 20.0 & 16.4 & 11.4 & 12.8
& 22.6 & 15.4 & 13.6 & 10.8
& \rulersecond{30.2} & \rulerbest{22.7}
& \rulersecond{19.2} & 0.4
& \rulerbest{31.8}
\\

\midrule

\multicolumn{30}{l}{\textit{Attention or hybrid models}} \\

\rowcolor{gray!30}
\rulermodel{Transformer}
& 96.6 & 90.8 & \rulerbest{47.5} & 0.0
& 92.7 & 81.6 & \rulerbest{48.1} & 0.0
& 37.8 & 8.6 & \rulersecond{18.1} & 0.0
& \rulersecond{47.6} & 35.7 & 11.5 & 0.0
& \rulersecond{34.8} & 27.8 & \rulerbest{16.4} & 0.2
& 41.6 & 35.0 & \rulerbest{29.4} & 0.2
& 12.2 & 0.6 & 9.2 & 0.0
& 29.4
\\

\rulermodel{Mamba-2}
& 61.2 & 47.6 & \rulersecond{44.1} & 0.0
& 55.9 & 44.1 & 26.1 & 0.0
& 28.0 & 13.2 & 1.3 & 0.0
& 29.9 & 27.1 & 2.2 & 0.0
& 27.6 & 24.4 & 12.0 & 0.2
& 38.8 & 29.2 & 21.0 & 0.0
& 38.8 & 25.6 & 0.0 & 0.0
& 21.4
\\

\rulermodel{Gated DeltaNet}
& 94.3 & 79.5 & 26.8 & 0.3
& 90.0 & 77.5 & 28.4 & 0.4
& 48.7 & 6.7 & 12.1 & 0.5
& 42.5 & 43.6 & 7.6 & 0.1
& 28.6 & 26.4 & 12.2 & 4.6
& 45.2 & 36.8 & 26.0 & 5.2
& \rulersecond{51.5} & \rulerbest{42.0}
& 0.6 & 0.1
& 29.9
\\

\rulermodel{KDA}
& 92.2 & 81.8 & 18.9 & 2.7
& 90.5 & 75.5 & 23.9 & 3.0
& 58.9 & 10.2 & 7.2 & 2.4
& 45.0 & 43.9 & 14.5 & 0.9
& 30.2 & 24.2 & \rulersecond{14.4} & 12.6
& 36.6 & 30.0 & 22.0 & 12.6
& 42.6 & \rulersecond{29.9}
& 0.8 & \rulersecond{0.7}
& 29.6
\\

\rulermodel{Mamba-3 (SISO)}
& \rulersecond{97.8} & \rulersecond{91.0}
& 39.0 & 0.0
& \rulerbest{96.5} & 82.7 & 36.3 & 0.0
& 48.3 & 9.2 & 9.8 & 4.5
& 44.1 & \rulersecond{46.7} & 22.1 & 6.5
& 34.0 & \rulerbest{34.2} & 11.6 & 0.0
& 43.0 & 37.0 & 26.0 & 0.4
& 16.0 & 19.8 & 13.8 & 0.0
& 31.1
\\

\rulermodel{Mamba-3 (MIMO)}
& 96.2 & 89.8 & 26.6 & 0.0
& \rulersecond{95.5} & \rulerbest{85.8}
& 32.0 & 0.0
& 54.1 & \rulerbest{27.4}
& 14.5 & \rulerbest{14.8}
& \rulerbest{53.5} & \rulerbest{49.7}
& \rulersecond{30.8} & 5.9
& 33.2 & 27.6 & 11.4 & 3.6
& 39.6 & 32.2 & 22.4 & 0.4
& 16.4 & 23.6 & \rulersecond{21.0}
& \rulerbest{3.4}
& 32.5
\\

\rulermodel{Gated DeltaNet-2}
& 94.0 & 77.3 & 20.3 & 8.7
& 84.5 & 57.0 & 20.9 & 11.2
& 51.7 & 4.9 & 12.3 & 0.8
& 47.0 & 45.3 & 16.1 & 19.0
& 24.6 & 24.8 & 14.2 & \rulersecond{15.4}
& 42.8 & 37.4 & 23.2 & \rulerbest{17.0}
& 13.8 & 12.2 & 1.4 & 0.4
& 28.5
\\

\rulermodel{OJA}
& 87.3 & 69.2 & 25.6 & \rulerbest{11.3}
& 86.0 & 49.8 & 28.4 & \rulersecond{13.1}
& \rulersecond{60.1} & 5.8 & 16.1 & 4.5
& 36.6 & 36.8 & 15.1 & 9.8
& 32.6 & \rulersecond{30.2} & 12.8 & \rulerbest{17.8}
& 42.8 & 40.0 & 23.4 & \rulersecond{16.8}
& 5.2 & 1.0 & 0.0 & 0.0
& 27.8
\\

\rulermodel{OJA2}
& 94.5 & 82.1 & 27.1 & 8.2
& 84.4 & 66.4 & 31.7 & 11.5
& 57.1 & 8.6 & 17.1 & 2.5
& 28.9 & 32.9 & 25.7 & \rulersecond{26.7}
& 29.4 & 20.8 & 11.6 & 14.0
& 39.6 & 31.2 & 20.2 & 15.0
& 8.9 & 4.6 & 0.0 & 0.0
& 28.6
\\

\rulermodel{Gated Slot Attention}
& \rulerbest{98.8} & \rulerbest{95.9}
& 38.7 & 0.1
& 90.9 & \rulerbest{87.8} & 41.1 & 0.1
& 45.1 & 6.4 & 1.9 & \rulersecond{11.1}
& 41.2 & 45.6 & 3.7 & 3.1
& \rulersecond{34.8} & \rulerbest{34.2}
& 13.8 & 0.0
& \rulerbest{48.4} & \rulerbest{45.4}
& \rulersecond{28.6} & 1.2
& \rulerbest{53.4} & 28.5
& \rulerbest{25.2} & 0.0
& \rulersecond{33.0}
\\

\rowcolor{blue!10}
\rulermodel{Gated Slot Attention-2}
& 96.0 & 87.2 & 41.3 & \rulersecond{10.6}
& 91.7 & 83.0
& \rulersecond{44.1} & \rulerbest{17.1}
& \rulerbest{68.5} & \rulersecond{25.6}
& \rulerbest{19.0} & 7.1
& 45.9 & 42.7
& \rulerbest{34.1} & \rulerbest{31.3}
& \rulerbest{35.6} & 25.6 & 12.6 & 14.4
& \rulersecond{47.6} & \rulersecond{41.0}
& 24.6 & 16.2
& 42.6 & 6.8 & 0.8 & 0.2
& \rulerbest{36.2}
\\

\bottomrule

\end{tabular}
\end{adjustbox}

\caption{
Accuracy on the remaining RULER tasks across context lengths.
Avg. is computed over all 28 settings.
All non-Transformer models in the attention/hybrid group use SWA.
Best and second-best results within each model family are bolded and underlined, respectively.
}
\label{tab:ruler_remaining}

\end{table*}

\end{document}